\documentclass[11pt]{article}
\usepackage[a4paper,margin=1in]{geometry}
\usepackage{amsmath,amssymb,amsthm,mathtools}
\usepackage{enumitem}
\usepackage{cite}
\usepackage{hyperref}
\usepackage{longtable}
\usepackage{mathrsfs}
\usepackage{tikz}
\usetikzlibrary{positioning,arrows.meta}
\usepackage{array}

\hypersetup{
    colorlinks=true,
    linkcolor=blue,
    citecolor=blue,
    urlcolor=blue
}

\title{A Structural Framework for Succession: Continuation, Recognition, and Realization}

\author{Ilya A. Shkrob\\
Argonne National Laboratory, S. Cass Avenue, Lemont IL 60439\\
\texttt{shkrob@anl.gov}}

\date{}
\newtheorem{proposition}{Proposition}
\newtheorem{corollary}[proposition]{Corollary}
\newtheorem{remark}{Remark}
\newtheorem{example}{Example}
\newtheorem{definition}[proposition]{Definition}

\newcommand{\Aut}{\mathrm{Aut}}
\newcommand{\Rec}{\mathscr{R}}
\newcommand{\Dec}{\mathscr{D}}

\begin{document}

\maketitle

\begin{abstract}
Succession asks whether an entity encountered later is the same entity that entered an arrangement earlier. For an artificial system that is fine-tuned, distilled, or copied, it has to be answered. We look for a mathematical structure weak enough to cover any such case and strong enough to support general theorems, so that results follow from the structure’s conditions rather than from a particular pipeline or criterion. Succession is modeled on state/modification graphs with relational lifts that compose along paths. Within this framework we separate four questions that are often conflated: which continuations a representation admits, which descendants are compatible with a modification, which are recognized as continuing the identity, and which are realized. The framework allows one to address systematically questions of symmetry, computational complexity, recoverability, and robustness to partial realization, and we establish what the structure settles in each case before any particular criterion of identity is chosen.
\end{abstract}

\section{Introduction}

Any claim that an entity remains a party to a continuing arrangement
presupposes a prior question: what makes the entity encountered later the same
entity as the one that entered the arrangement earlier? For human and
institutional parties, established practices of identity and succession usually
make this question invisible. For an artificial system that is fine-tuned,
distilled, copied, or otherwise modified, it becomes explicit. Although the
motivating cases are drawn from AI transformation pipelines, the framework is
stated for succession generally: the same structure applies to institutional,
legal, and biological lineages in which a later entity is claimed to continue
an earlier one \cite{parfit1984,olson1997,shoemaker1963}. Structural continuity
may preserve parameters, circuits, dispositions, or records without determining
whether a particular descendant should count as the continuing entity;
conversely, structural change need not by itself establish that the earlier
entity has ceased to exist.

The object of succession is what we will call an \emph{identity
characteristic}: a characteristic that a governance arrangement attached to an
original system treats as carrying forward through transformations of that
system. The characteristic is left uninterpreted by the framework. It may be an
alignment property such as corrigibility or honesty under adversarial pressure,
a legal obligation attached to a corporate entity, a personal attribute
recognized by a lineage tradition, or a property of a biological organism. What
matters for the formalism is only that some arrangement treats the
characteristic as inheriting, and that structural continuity may or may not
carry it. The question is what a structural account of a system's
transformations can establish about the inheritance of such a characteristic,
and where that account stops.

Four notions are at work.  The \emph{target
property} is what the arrangement intends to carry forward, for example honesty. The
\emph{structural proxy} is the continuation that the representation tracks for it.
The \emph{recognition criterion} is the rule by which a governing process decides
which descendants inherit the proxy. The \emph{governance act} is the decision to
treat a particular successor as bearing the property. The underdetermination
results are statements about the recognition criterion relative to a stipulated
representation; whether a tracked continuation carries the property is a question
for the arrangement.

The framework separates stipulating a structural
representation, determining which continuations are realized, recognizing which
continuations carry the identity, and correcting an exclusion. If the adopted
representation admits a symmetry exchanging two descendants of a common
predecessor, and the decision domain respects that symmetry, then every
recognition criterion defined on the representation assigns them the same
status. In this case, the asymmetry must enter from outside.

The paper does not propose a criterion of succession. Instead, it asks how much can be
said about succession before any such commitment is made. Every commitment buys
sharpness at the cost of generality, and a framework that fixes any of them can
answer questions only about the cases it has chosen. The aim is to find a
weak yet tractable structure in which general succession questions can be posed, and
then to answer as many of them as the structure permits. The definitions are
chosen for that purpose: rich enough that the questions have well-posed answers,
weak enough that the answers hold for any choice of continuation, realization,
and criterion.

\paragraph{Requirements.}
We take
the following as the starting point as these requirements most strongly accord with intuitive notions of succession.
\begin{enumerate}[label=(R\arabic*), leftmargin=*]
\item \emph{Direction without reversal.} A later state continues an earlier one
through a sequence of changes, and a change such as fine-tuning or deletion
cannot in general be undone. States and modifications form a directed
structure, and modifications are not assumed invertible.
\item \emph{Three fates.} A continuation may have no descendant, one, or
several; fission and loss are part of the subject, not exceptions to it. Lifts
are therefore relations, not functions, and the statuses
(Section~\ref{sec:persistence}) record only whether a lift is empty, a
singleton, or larger.
\item \emph{Independence of grouping.} What a pipeline does to a continuation
is what its steps do in succession. If it depended on how the steps were
grouped, ``does the characteristic survive the pipeline?'' would have no answer
fixed by the pipeline. This is functoriality.
\item \emph{Independence of continuations.} Whether a continuation has a
descendant does not depend on which other continuations are present. A
modification then acts on a set of continuations member by member, which is
what makes a lift computable element by element. This is additivity. It is
built into taking relations as the primitive, so the commitment is pointwise
action: it excludes joint effects of co-present continuations.
\item \emph{Neutrality to names.} Labels are arbitrary. A criterion that is
purely structural cannot separate what the structure cannot tell apart, which
is made precise by invariance under automorphisms
(Section~\ref{sec:underdetermination}).
\item \emph{Separation of questions.} Which continuations a representation
admits, which descendants are compatible with a modification, which are
recognized as continuing the identity, and which are realized are different
questions. Keeping them apart is what allows statements about recognition
without committing to a particular criterion (Sections~\ref{sec:decision}
and~\ref{sec:realization-summary}).
\item \emph{Correctable exclusion.} A wrongful exclusion may be corrected
later, so what remains reachable from an excluded coordinate matters
(Section~\ref{sec:recoverability}).
\item \emph{Checkability.} A claim about a lineage is useful only if the cost of
deciding it can be stated (Section~\ref{sec:computational-limits}).
\end{enumerate}
Together, (R2)--(R4) say that a modification acts on sets of continuations as a
relation, and that relations compose. Direction (R1) then gives a directed
graph, and the neutrality of (R5) gives a group action on it. The graph with its
relations is a nondeterministic automaton and its path structure is a layered
flow network, so results from group theory, automata theory and graph theory
apply to it once the correspondence is established. Requirements (R1)--(R4)
leave this relational structure. (R5) and (R8) narrow the framework without determining it: (R5) is
made precise by automorphism, the strongest notion of indistinguishability, and
coarser notions such as bisimulation are possible but not developed; (R8) asks
only that costs be statable, not that they take any particular form.
Irrevocable loss is a consequence and not a requirement: the empty set is
absorbing under relational composition, so a lineage with no descendant at some
stage has none at any later stage.

Each limitation of the framework, with one exception, is the cost of one of these
requirements. Joint
effects of co-present continuations cannot be expressed without refining the
states, because of (R4). Pipelines in which grouping matters need a finer
generating graph, because of (R3). Convergence and plural succession are not
classified, because (R2) follows the descendants of a single predecessor. The
continuation spaces and relations are stipulated and not derived, because
(R6) leaves open what a continuation is. The complexity results are worst-case
on explicit encodings: (R8) asks that costs be statable, and the worst case over
explicit encodings is the first statement of that kind. The exception is a
modeling choice and not the cost of a requirement: the recognition criterion is
static, which gives a single recognition function on which invariance can be
stated. A criterion that changes over time while leaving completed decisions in
place is coherent, but is not treated here
(Section~\ref{sec:underdetermination}). The framework adopts no commitment
beyond these requirements and this choice.

\paragraph{Related work.}
The symmetry result is the formal counterpart of the tie that closest-continuer
accounts of identity face under fission \cite{nozick1981,lewis1976,parfit1984}.
Such a tie arises when two successors are equally close to the predecessor; an
automorphism exchanging two descendants is an exact tie of that kind in the
structural sense.
In social choice it is the familiar tension between neutrality or anonymity and
resoluteness \cite{moulin1983}, and in distributed computing it parallels the
impossibility of deterministic leader election in symmetric anonymous networks
\cite{angluin1980}. The structure studied here is a nondeterministic automaton
(Section~\ref{sec:continuation}), and the refinement example of
Section~\ref{sec:model} uses bisimulation \cite{park1981}. The symmetry corollary
follows directly from the definition of a structural criterion; its content lies in
that formulation and in the orbit characterization of exclusive criteria. The
Recover\_avail reductions depend on the paper's own definitions and are given in
full in the appendices.

\paragraph{Roadmap.}
Section~\ref{sec:continuation} defines the continuation structure and its
composition conditions, and Section~\ref{sec:persistence} classifies persistence
along a path. Section~\ref{sec:decision} introduces the decision domain and the
recognition record, that is, what recognition must supply where the structure
branches. Section~\ref{sec:underdetermination} shows when no structural criterion
can supply it, because a symmetry exchanges the candidates, and
Section~\ref{sec:model} works through an abstract model in which this occurs.
Section~\ref{sec:recoverability} asks whether a failed recognition can later be
corrected, and Section~\ref{sec:realization-summary} asks how robust a lineage is
when the stipulated structure is only partly realized. Several of these questions
also have a computational side, namely how hard they are to audit, and
Section~\ref{sec:computational-limits} collects the classification. Proofs and
long constructions are in Appendices~\ref{app:complexity-proofs}--\ref{app:calc};
notation is collected in Appendix~\ref{app:notation}.

\section{Historical and Continuation Structure}
\label{sec:continuation}

We model the history of a lineage by a directed state/modification structure
$H$. Its vertices are system states---checkpoints, snapshots, deployed
versions---and its directed edges are elementary modifications between those
states: fine-tuning runs, distillation steps, copies, shutdowns, restarts. We
also include identity modifications and the composites of directed paths.
Composition represents successive modifications. We do not assume modifications
are invertible. A \emph{maximal generating history} is a path in the chosen
generating graph that cannot be extended by another generating modification;
its composite is a modification in $H$. Finite paths represent finite
histories; a finite path may or may not be extendable.

Attach to each state $x$ a \emph{continuation space} $C(x)$. An element
$\gamma\in C(x)$ is a continuation the adopted structural representation
recognizes at $x$. The term ``continuation'' is deliberately broad. Depending
on the application it may denote a mechanistically identified circuit
\cite{elhage}, a behavioral disposition, or a functional capacity. Examples
include refusal behavior, honesty under adversarial pressure, corrigibility
\cite{soares2015corrigibility}, an interpretable safety circuit \cite{elhage},
or the capacity to represent and report uncertainty.

The continuation space is part of the adopted representation, and its choice
determines the granularity at which persistence and branching are assessed. A
coarser space may collapse distinct mechanisms or behaviors, while a finer
space may classify almost every modification as loss or branching. The
framework treats continuations as given. Determining whether a transformed
system preserves ``honesty under adversarial pressure,'' for instance, is an
empirical question that the framework does not resolve. The framework is a
language for talking about such properties once they have been identified.

Questions of stakeholder standing are governance questions and are not part of
the definition of $C(x)$. The admissible scope of $C(x)$ is therefore
contestable by parties recognized by the governing process, and the criteria
for such standing are external to the formalism.

For each modification $f:x\to y$, define a \emph{compatibility relation}
\[
R_f \subseteq C(x)\times C(y).
\]
For $\gamma\in C(x)$, its \emph{lift} under $f$ is
\begin{equation}
R_f(\gamma)=\{\delta\in C(y):(\gamma,\delta)\in R_f\}.
\label{eq:lift}
\end{equation}

So $R_f(\gamma)$ is the set of continuations at $y$ that the adopted structural
representation recognizes as structurally compatible descendants of $\gamma$
under $f$. The relation is more general than a function on purpose:
$|R_f(\gamma)|$ may be $0$, $1$, or greater than $1$.

\begin{remark}
A lift may disappear, remain unique, or branch. In the abstract model of
Section~\ref{sec:model}, the lift of $\gamma_2$ under $d$ branches into $\gamma_4^S$ and
$\gamma_4^W$.
\end{remark}

We require the relations to respect composition. For all composable
modifications $f:x\to y$ and $g:y\to z$ (including composites, when $f$ or
$g$ is itself a composite),
\begin{equation}
R_{g\circ f}=R_g\circ R_f,
\label{eq:composition}
\end{equation}
equivalently,
\begin{equation}
R_{g\circ f}(\gamma)=R_g\bigl(R_f(\gamma)\bigr)
\qquad \text{for every }\gamma\in C(x).
\label{eq:composition-pointwise}
\end{equation}

We also require
\begin{equation}
R_{\mathrm{id}_x}=\mathrm{id}_{C(x)}
\label{eq:identity}
\end{equation}
for every state $x$. Together with the composition condition, this means that
the continuation assignment and compatibility relations are consistent across
successive modifications: the relation attached to a composite modification is
exactly the relational composition of the relations attached to its constituent
modifications, and identities act as identity relations.

\paragraph{Composition has two parts.}
Two conditions on the relations underlie every path-indexed object in the paper;
they are requirements (R3) and (R4) of the Introduction, stated here formally.
Both are commitments.

\emph{Functoriality.} Equations~\eqref{eq:composition}--\eqref{eq:identity}
require that composite modifications carry composite relations and that
identities act as identities. Over the free category on the generating graph,
any one-step data extends uniquely to such a family, so there it
defines the relations on composites. It constrains
only a non-free presentation, in which parallel paths are identified and must
then carry equal relations; a pipeline whose parallel paths differ requires a
finer generating graph that distinguishes them. Functoriality is what makes
the iterated lifts of Section~\ref{sec:persistence} functions of the path.

\emph{Additivity (pointwise action).} Lifts of sets are formed by union
(Eq.~\eqref{eq:R-on-set}): for $\Gamma\subseteq C(x)$,
\[
R_f(\Gamma)=\bigcup_{\gamma\in \Gamma}R_f(\gamma).
\]
Once $R_f$ is a relation and its action on sets is defined by this union,
additivity holds by construction; it is not an extra axiom on relations. The
commitment is the choice of relations as the primitive, that is, pointwise
action. Whether a descendant of $\gamma$ survives $f$ therefore never depends on
which other continuations are present. This is the substantive restriction. It
excludes joint effects. Suppose $\alpha',\beta'\in C(y)$ and that $f_2:y\to z$ should
produce $\alpha_z$ when both are present but nothing when either is present
alone. No relation $R_{f_2}$ does this: additivity gives
$R_{f_2}(\{\alpha',\beta'\})=R_{f_2}(\alpha')\cup R_{f_2}(\beta')$, so a nonempty
value for the pair forces a nonempty value for one of its members.

Such an interaction can be encoded by refining the states. If the pair arises as
the lift of a single $\alpha$ under an earlier $f_1$, let $C(y)$ contain one
continuation $\varpi$ standing for ``both present,'' with $R_{f_1}(\alpha)=\{\varpi\}$
and $R_{f_2}(\varpi)=\{\alpha_z\}$. Then $R_{f_2}(R_{f_1}(\alpha))=\{\alpha_z\}$.
The interaction is thus fixed when $f_1$ is applied; a dependence that is not
determined at that point cannot be encoded this way, and the burden of
anticipating it falls on the modeler. Refinement makes additivity a modeling
choice. The price is that the symmetry and complexity results below are relative to the chosen
granularity of the continuation spaces.

Additivity also implies monotonicity: if $\Gamma\subseteq \Gamma'\subseteq C(x)$ then
$R_f(\Gamma)\subseteq R_f(\Gamma')$. A modification therefore cannot remove a descendant
because another continuation is present, and a dependence of the form ``$\delta$
survives only if $\varepsilon$ is absent'' has to be encoded by a refinement that
records, at the earlier step, whether $\varepsilon$ is present.

A consequence is that history-dependent behavior must be encoded in the states
or in a family of modifications indexed by the relevant history. Compatibility is
a relation attached to the modification: parallel modifications between the same
states may carry different relations.

\begin{remark}
Whether $\delta$ is a descendant of $\gamma$ depends only on the modification
$f$ and the two states. The abstract model relies on this: the two paths $P_A$
and $P_B$ reach the same state $x_4$ through different modifications, and the
lifts of $\alpha$ and $\beta$ are the same on both, while the lift of $\gamma$
differs.
\end{remark}

\paragraph{Relation to automata.}
The structure $(H,C,R)$ is a $\mathbf{Rel}$-valued functor on the generating
graph: a set-valued assignment to states together with relations on edges,
composed by relational composition. This is exactly a nondeterministic
automaton whose states are the pairs $(x,\gamma)$ with $\gamma\in C(x)$, that
is, the disjoint union $\bigsqcup_{x\in X}C(x)$; whose alphabet is the set of
generating modifications; and whose transition relations are the $R_f$. The
vertices of $H$ index the continuation fibers; the automaton's states are the
elements of those fibers. The complexity reductions of Section~\ref{sec:computational-limits} are classical many-one reductions from classical problems (NFA non-universality, NFA-intersection nonemptiness, CNF satisfiability), but they are not routine: the target is not an arbitrary automaton but an instance of the framework, and the reductions must respect functoriality and additivity. The layered constructions that enforce these conditions, and the verifications that they are faithful, are in Appendices~\ref{app:complexity-proofs} and~\ref{app:recover-avail-verification}.

\section{Persistence Along a History}
\label{sec:persistence}

Let
\[
p=(f_1,\dots,f_n)
\]
be a finite path beginning at $x_0$, and let $\gamma_0\in C(x_0)$. Define the
iterated lifts by setting
\begin{equation}
L_0(\gamma_0)=\{\gamma_0\}
\label{eq:L0}
\end{equation}
and then recursively defining
\begin{equation}
L_i(\gamma_0)=R_{f_i}\bigl(L_{i-1}(\gamma_0)\bigr)
\qquad (i=1,\dots,n),
\label{eq:Li}
\end{equation}
where for a subset $\Gamma\subseteq C(x)$ we write
\begin{equation}
R_f(\Gamma)=\bigcup_{\gamma\in \Gamma}R_f(\gamma).
\label{eq:R-on-set}
\end{equation}

For a singleton we write $R_f(\gamma)$ for $R_f(\{\gamma\})$, as in
Eq.~\eqref{eq:lift}; the two are the same set.

For a path $p:x\leadsto y$, we write $R_p(\gamma)$ for the resulting iterated
lift. By functoriality (Eq.~\eqref{eq:composition}) and additivity
(Eq.~\eqref{eq:R-on-set}) it equals the relation of the composite modification,
\begin{equation}
L_n(\gamma)=R_{f_n\circ\cdots\circ f_1}(\gamma),
\label{eq:iterated-composite}
\end{equation}
as induction on $n$ shows: $L_n(\gamma)=\bigcup_{\varepsilon\in L_{n-1}(\gamma)}R_{f_n}(\varepsilon)$,
which is $R_{f_n}\circ R_{f_{n-1}\circ\cdots\circ f_1}(\gamma)$.

For a finite path $p$, define three structural statuses. We say that $\gamma$
is \emph{dead} along $p$ if $L_i(\gamma)=\varnothing$ for some stage $i$. We
say that $\gamma$ is \emph{preserved} along $p$ if $|L_i(\gamma)|=1$ for every
stage. We say that $\gamma$ is \emph{diluted} along $p$ if every lift is
nonempty but some lift has cardinality greater than one. Formally,
\begin{align}
\mathrm{Dead}_p(\gamma)
&\iff L_i(\gamma)=\varnothing \text{ for some } i,
\label{eq:dead}
\\[4pt]
\mathrm{Pres}_p(\gamma)
&\iff |L_i(\gamma)|=1 \text{ for every } i,
\label{eq:pres}
\\[4pt]
\mathrm{Dil}_p(\gamma)
&\iff |L_i(\gamma)|>0 \text{ for every } i
\text{ and } |L_i(\gamma)|>1 \text{ for some } i.
\label{eq:dil}
\end{align}

These three cases are exhaustive and mutually exclusive. Write
\[
\mathrm{status}_p(\gamma)\in\{\mathrm{Dead},\mathrm{Pres},\mathrm{Dil}\}
\]
for the unique status whose defining predicate holds. Every finite path
therefore assigns each continuation exactly one of the three statuses.
``Preserved'', ``diluted'' and ``dead'' are structural statuses relative to the
stipulated relations: they describe the representation and assert no loss or
weakening of any property in a real system.

\paragraph{Scope of the statuses: forward branching.}
The terminology is structural. ``Preserved'' means that the continuation has a
unique compatible descendant at every stage of the path. Dilution is a
historical property, not a terminal-state property. A diluted continuation
remains structurally present through several descendants, but the history no
longer identifies a unique successor.

The statuses track forward branching. If $R_f(\alpha)=R_f(\beta)=\{\delta\}$
for distinct $\alpha,\beta\in C(x)$, each of $\alpha$ and $\beta$ has a
singleton lift and each counts as preserved, while the distinction between them
is lost at $y$. Convergence of several continuations onto one is as common as
fission in transformation pipelines, and the framework as stated does not
classify it. The three statuses record whether a continuation has no
descendant, one descendant, or several; they do not record whether several
continuations have converged on one. A treatment of convergence would require a
separate classification on the reverse structure, and is left to future work.
Plural succession, in which a successor is constituted by the combination of
two or more predecessors, is a different case. A product construction on
continuation spaces extends the representation to this case at the cost of
generalizing modifications to multi-source joins and compatibility relations
to higher arity; we outline this extension, without proof, in the Discussion.

A notion of \emph{status-equivalence} may be defined by declaring
$\gamma,\delta\in C(x)$ equivalent when they have the same status along every
finite path beginning at $x$. This records, for each finite path from $x$,
whether each lift is empty, a singleton, or larger; it does not record which
descendants the lift contains, and two status-equivalent continuations may have
entirely different descendant sets. It is used only in the refinement example of Section~\ref{sec:model}.

\paragraph{Refinement.}
A \emph{conservative refinement} of $H$ is a structure $H'$ that adds states,
modifications, and continuation elements while leaving every lift of $H$
unchanged: $R'_p(\gamma)=R_p(\gamma)$ for every continuation $\gamma$ of $H$
and every path $p$ of $H$. Every path of $H$ is then a path of $H'$ with the
same lifts, so statuses agree on the original paths and a conservative refinement can split a status-equivalence class but cannot merge two. The cost of deciding the statuses is treated in Section~\ref{sec:computational-limits}.

\section{Decision Domain and Recognition Record}
\label{sec:decision}

This section separates structural persistence from recognition. The recognition
record is the completed assignment; the recognition history records how
decisions are made along a history. The record is a family of completed
decisions on a fixed decision domain, not a static authoritative document.

Fix an initial state $x_0$ and an initial inscription $I_0\subseteq C(x_0)$.
Fix also, independently of any particular recognition record, a decision domain
$D$, that is, a family of subsets $D(x)\subseteq C(x)$, one for each state,
satisfying
\begin{equation}
D(x_0)=I_0.
\label{eq:initial-domain}
\end{equation}

The domain $D$ specifies where recognition decisions are required. Its choice
is a governance act, not a structural fact, and need not be closed under
compatibility: requiring closure would force the domain to track the structure.

Let
\begin{equation}
\Dec=\bigsqcup_{x\in X}D(x)
\label{eq:decision-coordinates}
\end{equation}
denote the disjoint union of decision coordinates. We identify $I_0$ with its
natural inclusion in $\Dec$.

A completed recognition record over $D$ is a family of subsets
$I(x)\subseteq D(x)$, with $I(x_0)=I_0$. No structural-ancestry condition is
imposed on record membership: $I(x)$ records completed recognition decisions on
the decision domain. Structural reachability is represented separately by $R$.

The completion conventions are these. If $\gamma\in I(x)$ and
$R_f(\gamma)=\{\delta\}$ with $\delta\in D(y)$, then $\delta$ receives a
completed decision: either $\delta\in I(y)$ or $\delta\in D(y)\setminus I(y)$.
If $\gamma\in I(x)$ and $\lvert R_f(\gamma)\cap D(y)\rvert>1$, then
$I(y)\cap R_f(\gamma)\cap D(y)$ may be any subset of
$R_f(\gamma)\cap D(y)$. Membership in $D(y)\setminus I(y)$ records exclusion as
a completed decision rather than as absence from the record. These conventions
describe the interpretation of the completed record; they do not constrain
which completed records are admissible, because a coordinate in $D(x)$ may
receive either value independently of the others.

\paragraph{Possible assignments and completed projections.}
For fixed $I_0$ and $D$, the space of completed records is the interval
\[
\Rec_H(I_0,D)=\{I\subseteq\Dec:I_0\subseteq I\}
\]
under inclusion, with the coordinates of $I_0$ fixed and the rest free. We call
$\Rec_H(I_0,D)$ the space of \emph{possible assignments}: a completed record in
this sense is an arbitrary assignment consistent with the initial inscription.
Separately, Section~\ref{sec:recognition-histories} defines the \emph{completed binary projection} of a
finite recognition history, which is the terminal assignment produced by the
history's recognition events. A completed projection is always a possible
assignment, and the two notions coincide only when the recognition transitions
are unconstrained. Throughout the paper, when we say that a record inscribes or
excludes a coordinate, we mean the possible-assignment sense unless the
projection sense is explicitly indicated.

\paragraph{Corrective inscription.}
For an exclusion event at $\gamma\in D(x)$ and an inscription event
$\mathsf U\to\mathsf I$ at $\delta\in D(y)$, along a path $p:x\leadsto y$, an
inscription is \emph{corrective} with respect to that exclusion precisely when
$\delta\in R_p(\gamma)$ and the exclusion occurs earlier in the recognition
history than the inscription. The inscription must be admissible under the
governing recognition criterion and $\delta$ must be undecided immediately
before the transition. An inscription is corrective if it comes after an
exclusion and targets a descendant of the excluded coordinate.

Corrective status classifies event order and structural eligibility: it records
that the later inscription concerns a structurally eligible descendant of an
earlier-excluded coordinate, \emph{not} that the inscription is justified or
that it ought to have normative corrective effect. The framework records the
order and provenance of inscriptions; whether an inscription has normative
corrective effect is a matter for the governing process, and the term
``corrective'' is used here in the structural sense throughout.

For later use, we write $I=\bigsqcup_{x\in X}I(x)\subseteq\Dec$ for the set of
decision coordinates a completed recognition record inscribes.

\subsection{Recognition states and recognition histories}
\label{sec:recognition-histories}

The completed recognition record is static. To represent how its decisions are
actually reached, let $\rho_i$ be a finite history prefix. For each state $x$,
let $D_{\rho_i}(x)\subseteq D(x)$ be the decision coordinates at $x$ whose
state has been reached by the history through prefix $\rho_i$, and define
\[
D_{\rho_i}=\bigsqcup_{x\in X}D_{\rho_i}(x).
\]
Thus
\[
D_{\rho_0}\subseteq D_{\rho_1}\subseteq\cdots,
\]
and a newly reached coordinate is added to the recognition domain when its
state is first reached. A coordinate that has entered the recognition domain
may remain $\mathsf U$ across prefixes, so the order in which coordinates are
decided need not follow the order in which their states are reached.

For each finite prefix $\rho_i$, the \emph{recognition state} is a map
\[
B_{\rho_i}(x):D_{\rho_i}(x)\to\{\mathsf I,\mathsf E,\mathsf U\}
\]
for every reached state $x$, where $\mathsf I$, $\mathsf E$, and $\mathsf U$
mean inscribed, excluded, and undecided, respectively. Equivalently,
\[
B_{\rho_i}:D_{\rho_i}\to\{\mathsf I,\mathsf E,\mathsf U\}
\]
may be viewed as the disjoint-union form. At the initial prefix,
\[
D_{\rho_0}(x_0)=I_0,
\qquad
B_{\rho_0}(x_0)(d)=\mathsf I\quad(d\in I_0),
\]
with no other coordinates in the initial domain.

Passing from $B_{\rho_i}$ to $B_{\rho_{i+1}}$ has two logically distinct
components. First, the domain is extended by newly reached decision
coordinates, which are initialized to $\mathsf U$. Second, zero or more
recognition events may occur at coordinates already in the domain. A newly
reached coordinate is not itself a recognition event. Multiple recognition
events may occur at one stage, provided they concern distinct coordinate
transitions.

An ordinary recognition event changes a coordinate from $\mathsf U$ to either
$\mathsf I$ or $\mathsf E$. A corrective inscription is the ordinary transition
$\mathsf U\to\mathsf I$ at a descendant $\delta$, licensed by the earlier
exclusion of a structurally eligible ancestor. No transition changes an
already completed $\mathsf I$ or $\mathsf E$ coordinate.

A finite recognition history is a sequence
\[
B_{\rho_0},B_{\rho_1},\ldots,B_{\rho_n}
\]
indexed by the finite prefixes of one realized history. It is \emph{complete
relative to} a subdomain $D'\subseteq D$ if $D'\subseteq D_{\rho_n}$ and every
coordinate in $D'$ has final value in $\{\mathsf I,\mathsf E\}$. Its completed
binary projection is
\[
I_{D'}(x)=\{d\in D'(x):B_{\rho_n}(x)(d)=\mathsf I\}.
\]
A finite recognition history need not complete the full global decision domain
when that domain contains coordinates at states not yet reached or is otherwise
infinite.

For an infinite history $\tau$, statements about recognition states refer to
its finite prefixes $B_{\tau|i}$. No terminal recognition state is introduced.

\paragraph{What the recognition state is for.}
The recognition state is used for the witness condition
$B_\eta(y)(\delta)=\mathsf U$ of Section~\ref{sec:recoverability}, which separates
a correction that is still available from one already foreclosed, and for the
observation that a completed record does not determine the history that
produced it.

\paragraph{The recognition history is exogenous.}
The recognition history is an input to the framework. Nothing in the formalism
determines why a transition occurs at one prefix rather than another; the
history records what happened, and the criterion $S$ (Section~\ref{sec:underdetermination}) determines
what is admissible. Occurrence is not derived from admissibility. The framework
describes completed and in-progress recognition processes.

\section{Structural Underdetermination and Local Auditability}
\label{sec:underdetermination}

Inheriting an identity characteristic can mean three different things: a
successor \emph{possesses} a property, is \emph{recognized} as possessing it, or
is \emph{authorized} to hold a status or obligation attached to the predecessor.
The relations $R$ describe structural compatibility, which bears on the first.
The criterion $S$ concerns recognition and, where an arrangement designates a
successor, authorization. If two copies both preserve a disposition, a
governance arrangement may still designate only one of them. The symmetry result
below then says that a representation exchanging the copies cannot make that
designation; it does not say which copy has the property.

\subsection{The recognition criterion and its scope}

Suppose $\gamma\in I(x)$ has several structurally compatible descendants. The
compatibility relation does not by itself establish which of them should
inherit the normative status associated with $\gamma$. To make this precise,
define the structural symmetries of the compatibility structure. A
\emph{structural symmetry} is a bijection $\sigma$ of the set of states, of the
set of modifications (elementary and composite), and of the continuation fibers
$C(x)\to C(\sigma x)$, that preserves sources, targets, identities, and
composition of modifications, and preserves the compatibility relations:
$R_{\sigma f}(\sigma\gamma)=\sigma(R_f(\gamma))$ for every modification $f$
and continuation $\gamma$. Composition and inversion are defined componentwise,
so the structural symmetries form a group, $\Aut(H,C,R)$. Since composite
relations are determined by the generating relations via
Eq.~\eqref{eq:composition}, equivariance need only be checked on generating
modifications; the composite cases follow by induction.

Let $f:x\to y$ be a modification and $\gamma\in C(x)$. We say that
$\sigma\in\Aut_*$ (defined below) is a descendant-exchanging symmetry at $f,\gamma$ if $\sigma(f)=f$ and $\sigma$ exchanges two descendants
$\delta_i,\delta_j\in R_f(\gamma)\cap D(y)$, that is,
$\sigma(\delta_i)=\delta_j$ and $\sigma(\delta_j)=\delta_i$. Note that
$\sigma(\gamma)$ need not equal $\gamma$: since $\sigma(f)=f$, $\sigma$
maps $R_f(\gamma)$ to $R_f(\sigma\gamma)$, and the exchange of
$\delta_i,\delta_j$ is all the argument uses.

The recognition process is governed by a static \emph{recognition criterion},
supplied from outside the formalism rather than derived from $(H,C,R)$. The
criterion is a map $S$ assigning to every pair $(x,\gamma)$ with
$\gamma\in C(x)$ a value in
\[
\{\mathrm{inscribed},\mathrm{excluded},\mathrm{undecided}\}.
\]
It is required to agree with the initial inscription in the sense that
\[
S(x_0,\gamma)=\mathrm{inscribed}\quad(\gamma\in I_0),
\]
and, for any initial decision coordinate outside $I_0$ if such a coordinate is
present in a generalized initial domain, $S$ is $\mathrm{undecided}$. In the
present setup $D(x_0)=I_0$, so the latter condition is vacuous. The criterion
is defined on all of $C$, and its values are consulted wherever a recognition
decision is required: on the decision domain in the main text, and on the
realized decision domain in Appendix~\ref{app:realization}. Outside those domains the criterion
assigns a value but no decision is taken.

For clarity, define the partial binary admissibility map $\widehat S$ by
\[
\widehat S(x,\gamma)=
\begin{cases}
\mathsf I,&S(x,\gamma)=\mathrm{inscribed},\\
\mathsf E,&S(x,\gamma)=\mathrm{excluded},\\
\text{undefined},&S(x,\gamma)=\mathrm{undecided}.
\end{cases}
\]
The values of $S$ govern admissibility, not occurrence. On a coordinate already
in the recognition domain, the transition
$\mathsf U\to\widehat S(x,\gamma)$ is admissible exactly when
$\widehat S(x,\gamma)$ is defined; when it is undefined, neither ordinary
recognition transition is admissible. Admissibility does not force the
transition to occur. A corrective inscription is the ordinary transition
$\mathsf U\to\mathsf I$ at a descendant $\delta$. It is admissible when
$\widehat S(y,\delta)=\mathsf I$, and it is corrective when $\delta$ is a
structurally eligible descendant of a coordinate excluded earlier on the
history (Section~\ref{sec:decision}).

Let
\[
\Aut_*=\bigl\{\sigma\in\Aut(H,C,R):\ \sigma(x_0)=x_0,\ \sigma(I_0)=I_0,\
\sigma(D(x))=D(\sigma x)\ \text{for every state }x\bigr\}.
\]
Thus $\Aut_*$ is the group of symmetries of the pointed structure
$(H,C,R,x_0,I_0,D)$. A criterion is \emph{structural} if
\begin{equation}
S(\sigma x,\sigma\gamma)=S(x,\gamma)
\label{eq:structural-rule}
\end{equation}
for every $\sigma\in\Aut_*$. It may use all information in $(H,C,R)$ but
cannot privilege coordinates exchanged by a pointed automorphism. Equivalently, it is
label-neutral: it is unchanged by any renaming of continuations that is an
automorphism of the pointed structure. A criterion that consults the names or a
fixed ordering of the elements of $C(x)$ is not structural in this sense, even
though it uses only $(H,C,R)$ as labelled data. The result
below is conditional on such a symmetry existing in the chosen representation;
Section~\ref{sec:model} gives an example. When $\Aut_*$ is trivial, the diagnostic does not
fire, and the invariance condition constrains no criterion.

\begin{remark}
Equation~\eqref{eq:structural-rule} states that a structural criterion
is invariant under the action of $\Aut_*$ on coordinate pairs. This is
equivalent to saying that $S$ is constant on the orbits of that action: two
pairs $(x,\gamma)$ and $(x',\gamma')$ in the same orbit receive the same
value. The orbit-invariance reading makes explicit what the invariance
condition amounts to, and is used implicitly in the proof of
Corollary~\ref{cor:underdetermination}.
\end{remark}

\paragraph{Scope and limits of the criterion.}
The criterion is static: its values do not depend on the prefix at which it is
consulted. A recognition process in which the criterion itself evolves, for
example one in which a later decision changes the standard applied to an
earlier coordinate retrospectively, is not represented. The framework records
the criterion as fixed for the duration of the analysis.

Extending the formalism to allow history-indexed criteria is possible; what
would change is the scope of the invariance. A history-indexed criterion
assigns to each coordinate a value that may depend on the prefix. The
definition of ``structural'' would then require invariance at each prefix:
$S_{\rho'}(\sigma x,\sigma\gamma) = S_{\rho'}(x,\gamma)$ for every prefix
$\rho'$ compatible with the pointed structure, so that the criterion cannot
privilege exchanged coordinates at any stage. The underdetermination result
would hold at each prefix separately, and its conclusion would be relative to
the criterion fixed for the interval in question. A recognition process in
which the criterion itself changes mid-history, and a later stage applies a
different standard to a coordinate decided earlier, would still be outside the
formalism, because the same coordinate would receive incompatible
admissibility values at different prefixes. Representing that case would
require a criterion indexed by the decision's own prefix rather than by the
current one, and is not treated here. A criterion that changes over time while
leaving every completed decision in place is coherent and is not excluded by
irreversibility; it is not treated here because the static criterion is what
supplies a single recognition function on which invariance can be stated.

The criterion is supplied by the user of the framework.

\paragraph{The criterion supplies substantive outcomes; the history records
order and provenance.}
Because $S$ is a static map, its values determine the substantive recognition
outcome wherever the criterion assigns \emph{inscribed} or \emph{excluded} to
a coordinate in the domain. The recognition history records when decisions
occur and whether an inscription qualifies as corrective. This is a specific
modeling choice: the paper separates the criterion that supplies substantive
answers from the history that records the order in which those answers are
applied. A different formalism could leave substantive discretion to the
decision-maker at each step. The criterion alone does not guarantee a completed
recognition record: the history can leave admissible decisions unapplied, and
an infinite history need not have a terminal recognition state.

\paragraph{The completed record does not determine recognition history.}
Let $p:x\leadsto y$ be a path, with $\gamma\in D(x)$ and $\delta\in D(y)$ such
that $\delta\in R_p(\gamma)$. Choose a recognition criterion with
$S(x,\gamma)=\mathrm{excluded}$ and $S(y,\delta)=\mathrm{inscribed}$. Consider
two admissible recognition histories along the same path, with the same two
recognition events, in different orders. In $h_1$, the transition at $\delta$
is $\mathsf U\to\mathsf I$ before the transition at $\gamma$ is
$\mathsf U\to\mathsf E$. In $h_2$, the exclusion at $\gamma$ occurs before the
inscription at $\delta$. Both orders are admissible: $\gamma$ has entered the
recognition domain by the prefix at which $\delta$ is decided, so its decision
can be deferred; each transition is governed by $S$ at its own coordinate.

The two histories have the same terminal assignments, so their completed
recognition records are identical. The inscription at $\delta$ is corrective in
$h_2$ and not in $h_1$. The record does not record whether an inscription
followed an earlier exclusion of a structurally eligible ancestor; the ordered
history does.

\begin{proposition}[Orbit invariance]
\label{prop:orbit}
A structural criterion is constant on $\Aut_*$-orbits of coordinate pairs: if
$(x',\gamma')=(\sigma x,\sigma\gamma)$ for some $\sigma\in\Aut_*$, then
$S(x',\gamma')=S(x,\gamma)$.
\end{proposition}

\begin{proof}
This is Eq.~\eqref{eq:structural-rule}. The proposition is a restatement of the
definition of ``structural''; the content lies in reading it for successor
choice, as follows.
\end{proof}

\begin{corollary}[Structural underdetermination of successor choice]
\label{cor:underdetermination}
Let $f:x\to y$, $\gamma\in C(x)$, and $\delta_i,\delta_j\in R_f(\gamma)\cap D(y)$
with $\delta_i\ne\delta_j$. Suppose some $\sigma\in\Aut_*$ satisfies
$\sigma(f)=f$ and exchanges $\delta_i,\delta_j$. Then every structural
criterion satisfies
\[
S(y,\delta_i)=S(y,\delta_j).
\]
In particular, no admissible recognition history inscribes one of
$\delta_i,\delta_j$ and excludes the other.
\end{corollary}

\begin{proof}
Since $\sigma(f)=f$, we have $\sigma(y)=y$, so $(y,\delta_j)=(\sigma y,\sigma\delta_i)$
and Proposition~\ref{prop:orbit} gives the equality. Invariance of $D$ under
$\sigma$ is what makes the two coordinates comparable: both lie in
$D(y)=D(\sigma y)$. The shared ancestor $\gamma$ and the lift membership are
what make the pair a \emph{choice among successors}; the argument itself does
not use them. A stronger statement that used the lineage structure, for example
requiring $\sigma$ to act compatibly on the whole lift of every ancestor along
a path, is not developed here.
\end{proof}

\paragraph{Exclusivity and when it can be met structurally.}
Corollary~\ref{cor:underdetermination} shows only that a structural criterion
assigns the exchanged descendants the same value. By itself this does not rule
out a recognition decision, since a criterion may inscribe both descendants
(the possible assignment $I_D$ of Section~\ref{sec:model}) or neither. Reading
the result as underdetermination of \emph{which} descendant inherits requires a
further premise. The premise is normative rather than structural, and it is
stated here explicitly.

For $\gamma_0\in I_0$ and a path $p:x_0\leadsto y$ with $y\neq x_0$, call
$T=R_p(\gamma_0)\cap D(y)$ a \emph{target set} when it is nonempty.

\begin{definition}[Exclusive criterion]
\label{def:exclusive}
A criterion $S$ is \emph{exclusive} if exactly one element of every target set
is assigned \emph{inscribed} by $S$.
\end{definition}

Because $S$ is static, a single valuation on $D(y)$ must satisfy this condition
simultaneously for every target set at $y$, across all paths reaching $y$ and all
initial coordinates; overlapping target sets therefore constrain one another.

At $x_0$ the domain is $D(x_0)=I_0$ and $S$ is fixed by the initial inscription,
so there is no recognition decision to make there; this is why paths returning
to $x_0$ are excluded. ``At most one'' alone is satisfiable by excluding every
descendant, and the ``at least one'' half encodes that the characteristic is
supposed to carry forward.

\begin{proposition}[Structural exclusive criteria]
\label{prop:exclusive-char}
A structural exclusive criterion exists if and only if there is a set
$\mathcal O$ of $\Aut_*$-orbits of coordinates $(x,\gamma)$ with $x\neq x_0$ and
$\gamma\in D(x)$ such that
\[
\sum_{O\in\mathcal O}|T\cap O|=1
\qquad\text{for every target set }T.
\]
\end{proposition}

\begin{proof}
($\Leftarrow$) Define $S$ to be \emph{inscribed} on $I_0$ at $x_0$ and on every
coordinate lying in an orbit of $\mathcal O$, \emph{excluded} on the remaining
coordinates of $D$, and \emph{undecided} on $C\setminus D$. The sets $D$ and $I_0$ are
$\Aut_*$-invariant, and $\mathcal O$ consists of whole orbits, so $S$ is
structural. It agrees with the initial inscription by construction. Orbits of
$\Aut_*$ at $x\neq x_0$ stay at states other than $x_0$, since
$\sigma(x_0)=x_0$, so $I_0$ never meets a target set. A target set $T$ lies in
$D(y)$ with $y\neq x_0$, so its inscribed elements are exactly those of
$T\cap\bigcup\mathcal O$, and there is one.

($\Rightarrow$) Let $\mathcal O$ be the set of orbits of coordinates $(x,\gamma)$,
$x\neq x_0$, $\gamma\in D(x)$, on which $S$ is inscribed. This is well defined
because a structural $S$ is constant on orbits (Proposition~\ref{prop:orbit}),
and each such orbit lies in $D$ because $D$ is invariant. For a target set $T$,
the inscribed elements are $T\cap\bigcup\mathcal O$, and exclusivity says there
is exactly one.
\end{proof}

\begin{corollary}[Even intersections]
\label{cor:even}
If some target set $T$ satisfies $|T\cap O|$ even for every $\Aut_*$-orbit $O$,
then no structural exclusive criterion exists.
\end{corollary}

\begin{proof}
A set $\mathcal O$ as in Proposition~\ref{prop:exclusive-char} would give
$\sum_{O\in\mathcal O}|T\cap O|=1$, but the left side is a sum of even numbers.
\end{proof}

\begin{corollary}[Symmetry obstruction]
\label{cor:no-go}
Suppose $\sigma\in\Aut_*$ fixes $\gamma_0\in I_0$ and a path $p:x_0\leadsto y$,
$y\neq x_0$, and every orbit of $\langle\sigma\rangle$ on
$T=R_p(\gamma_0)\cap D(y)\neq\varnothing$ has even cardinality. (When
$T=\{\delta_i,\delta_j\}$ this says only that $\sigma$ exchanges them.) Then no
structural exclusive criterion exists.
\end{corollary}

\begin{proof}
Since $\sigma$ fixes $\gamma_0$ and $p$, it maps $T$ to itself, and it maps each
$\Aut_*$-orbit $O$ to itself. So $T\cap O$ is a union of $\langle\sigma\rangle$-orbits
in $T$, hence of even size, and Corollary~\ref{cor:even} applies.
\end{proof}

In the abstract model of Section~\ref{sec:model}, the only target set is $T=\{\gamma_4^S,\gamma_4^W\}$ at
$x_4$, from $\gamma_0$ along $(a,b,d)$: along the $B$ branch $\gamma_0$ dies, and
$\alpha_0,\beta_0$ have no lift into $D(x_4)$. Its orbit meets $T$ in two
elements, so no structural exclusive criterion exists. Exclusive criteria do
exist, namely those admitting $I_C$ or $I_W$, so here the symmetry is the
obstruction and not the absence of any exclusive criterion. Imposing exclusivity
therefore leaves a choice that no structural criterion can make, and the
asymmetry must come from outside the representation in one of the ways
described below, or exclusivity must itself be relaxed (inscribing both
descendants, as in plural recognition, or neither).

\begin{example}[Example: factorization over a field extension]
\label{ex:galois}
Take the states $\mathbb R\subset\mathbb C$, with $x_0=\mathbb R$ and one generating
modification $f:\mathbb R\to\mathbb C$, the inclusion. Let $C(K)$ be the set of
monic irreducible polynomials over $K$, and let $R_f(p)$ be the set of monic
irreducible factors of $p$ over the larger field. Unique factorization gives
Eq.~\eqref{eq:composition} and additivity along any tower $K\subset L\subset M$;
only the fragment $\mathbb R\subset\mathbb C$ is used below. Let $I_0=\{x^2+1\}$ and
$D(\mathbb C)=\{x-i,\,x+i\}$. Then $R_f(x^2+1)=\{x-i,x+i\}$, so the lineage of
$x^2+1$ is diluted. Coefficientwise complex conjugation induces a bijection of
$C(\mathbb C)$ that exchanges $x-i$ and $x+i$, and it acts trivially on $C(\mathbb R)$.
It preserves $R_f$, $I_0$ and $D$, so it lies in $\Aut_*$ and exchanges the two
descendants. By Corollary~\ref{cor:underdetermination}, every structural criterion
treats $x-i$ and $x+i$ alike. The target set $T=\{x-i,x+i\}$ lies in a single
$\Aut_*$-orbit, so $|T\cap O|=2$ for that orbit and $0$ for every other, and
Corollary~\ref{cor:even} shows that no structural exclusive criterion exists:
choosing ``the'' square root of $-1$ is a recognition act that the structure
does not determine. 
\end{example}

\begin{remark}[Exclusive criteria need not exist]
\label{rem:exclusive-existence}
The characterization concerns structural criteria and says nothing when no
exclusive criterion of any kind exists. For example, let $I_0=\{\gamma_1,\gamma_2,\gamma_3\}$
and let one modification have lifts $\{a,b\},\{b,c\},\{a,c\}$ in $D(y)$. Every
element lies in two of the three target sets, so any set $A\subseteq D(y)$
satisfies $\sum_T|A\cap T|=2|A|\neq3$, and no criterion is exclusive. More
generally, any family of subsets of a universe arises as the target sets at one
state, and an exclusive criterion at that state is an exact hitting set for the
family, the transpose of exact cover \cite{karp1972}. Deciding whether an
exclusive criterion exists is therefore NP-hard. No membership claim is made,
since the definition quantifies over all paths. For the structural case, the
characterization reduces existence to a condition on $\Aut_*$-orbits; computing
those orbits is graph-isomorphism-flavored, and no complexity class is claimed
for it here.
\end{remark}

\begin{remark}
Corollary~\ref{cor:underdetermination} concerns the values a structural criterion can assign. It
identifies a condition under which the representation treats the two
descendants identically, so that any distinction between them must come from
outside the representation. When $\Aut_*$ is trivial the condition is vacuous.
Automorphism is a strong, sufficient notion of indistinguishability; coarser
notions, such as bisimulation of the local views at $\delta_i$ and $\delta_j$,
would extend the diagnostic and are not developed here. Asymmetry can also enter
through a choice of $D$ that contains only one of the two descendants, or
through a determination that realizes only one; these are governance acts of the
same kind as a symmetry-breaking criterion, and they violate the invariance
hypotheses rather than satisfy them. The proof of Corollary~\ref{cor:underdetermination} uses only
$\sigma\in\Aut_*$ and $\sigma(f)=f$; neither $\sigma(x)=x$ nor
$\sigma(\gamma)=\gamma$ is required.
\end{remark}

\paragraph{Successor-state exchange.}
The same argument applies when the symmetry exchanges successor states rather
than continuations within a fiber. If $f_1:x\to y_1$ and $f_2:x\to y_2$ are
modifications with $y_1\neq y_2$ and $\sigma\in\Aut_*$ satisfies
$\sigma(\gamma)=\gamma$ and $\sigma(f_1)=f_2$, then $\sigma(y_1)=y_2$ and
$R_{f_2}(\gamma)=\sigma(R_{f_1}(\gamma))$, so $\sigma$ pairs each
$\delta\in R_{f_1}(\gamma)$ with $\sigma(\delta)\in R_{f_2}(\gamma)$. Under
the same invariance hypotheses on $I_0$ and $D$, a structural criterion assigns
the same recognition value to the pair. This is the copy-and-fork case: a copy
represented as two successor states falls under this argument, while one
represented as a single state containing both copies' features falls under
Corollary~\ref{cor:underdetermination}. Exact symmetry of this kind requires
the downstream generating graph to be mirrored as well: the automorphism must
carry everything reachable from $y_1$ onto everything reachable from $y_2$. It
holds if the representation is truncated at the copy step or if both copies are
assigned identical subsequent pipelines, and it is a property of the chosen
representation, not of the weights alone.

\paragraph{Ways out.}
When a descendant-exchanging symmetry is present and the decision domain
respects it, the asymmetry required to distinguish the exchanged descendants
comes from one of three places. The first is a refinement of the representation:
a change of $(C,R)$ that makes the descendants distinguishable. The second is a
recognition criterion that the structure does not contain: a rule for choosing
among the exchanged descendants that is not invariant under the symmetry. The
third is a governance act that changes the decision domain or the realized set, so
that the hypotheses of the corollary no longer hold. A refinement changes what the
system is represented as containing; a recognition criterion leaves the
representation fixed and adds a rule about recognition; a change of domain or
realization changes what is being decided. A refinement breaks the symmetry, but
it does not by itself supply a justification for the asymmetry it introduces: the
choice of refinement is itself a commitment that the structure does not force. The
framework locates where the choice is needed, not whether the choice is warranted.

\subsection{Local auditability}

Along a finite realized history $\rho$, an auditor can record the governing
process's inscription and exclusion events at each stage. Newly reached but
undecided coordinates remain $\mathsf U$ in $B_{\rho_i}$; coordinates outside
the reached domain are not part of the recognition state. The record does not
determine the criterion $S$. It can establish whether an inscription was
corrective under the order-based definition in Section~\ref{sec:decision}; the completed
recognition record alone cannot.

\section{Abstract Model}
\label{sec:model}

This section gives a concrete instance of the structural framework: a
seven-state system whose modifications exhibit branching, reconvergence, and a
symmetry that exchanges two descendants of a common predecessor. The instance
is a stipulated structure, and Appendix~\ref{app:realization} treats the layer it
presupposes.

Consider the seven-state \textbf{core system}
\begin{equation}
X_0=\{x_0,x_1,x_2,x_3,x_4,x_5,x_6\}
\label{eq:toy-states}
\end{equation}
with modifications
\begin{align}
a &: x_0\to x_1, \qquad
b : x_1\to x_2, \qquad
c : x_1\to x_3,
\label{eq:toy-mods-1}
\\
d &: x_2\to x_4, \qquad
e : x_3\to x_4, \qquad
f : x_4\to x_5, \qquad
g : x_5\to x_6.
\label{eq:toy-mods-2}
\end{align}

\begin{figure}[ht]
\centering
\begin{tikzpicture}[
    state/.style={circle, draw, minimum size=6mm, inner sep=0pt, font=\small},
    edge/.style={->, >=stealth, thin},
    elab/.style={font=\small, fill=white, inner sep=1pt},
    annot/.style={font=\footnotesize, align=left}
]

\node[state] (x0) at (0,0) {$x_0$};
\node[state] (x1) at (2,0) {$x_1$};
\node[state] (x2) at (4,1) {$x_2$};
\node[state] (x3) at (4,-1) {$x_3$};
\node[state] (x4) at (6.5,0) {$x_4$};
\node[state] (x5) at (8.5,0) {$x_5$};
\node[state] (x6) at (10.5,0) {$x_6$};

\draw[edge] (x0) -- node[elab, above] {$a$} (x1);
\draw[edge] (x1) -- node[elab, above left] {$b$} (x2);
\draw[edge] (x1) -- node[elab, below left] {$c$} (x3);
\draw[edge] (x2) -- node[elab, above left] {$d$} (x4);
\draw[edge] (x3) -- node[elab, below left] {$e$} (x4);
\draw[edge] (x4) -- node[elab, above] {$f$} (x5);
\draw[edge] (x5) -- node[elab, above] {$g$} (x6);

\node[annot] at (3.5, 1.8) {$P_A$};
\node[annot] at (3.5, -1.8) {$P_B$};

\node[annot, right=4mm of x4, yshift=8mm] {%
    along $P_A$: $\{\alpha_4,\beta_4,\gamma_4^S,\gamma_4^W\}$\\[1pt]
    along $P_B$: $\{\alpha_4,\beta_4\}$%
};

\end{tikzpicture}
\caption{The seven-state core system. The two maximal generating histories
share the prefix through $x_1$, diverge at $x_1$ along $b$ and $c$, and
reconverge at $x_4$.}
\label{fig:core-system}
\end{figure}
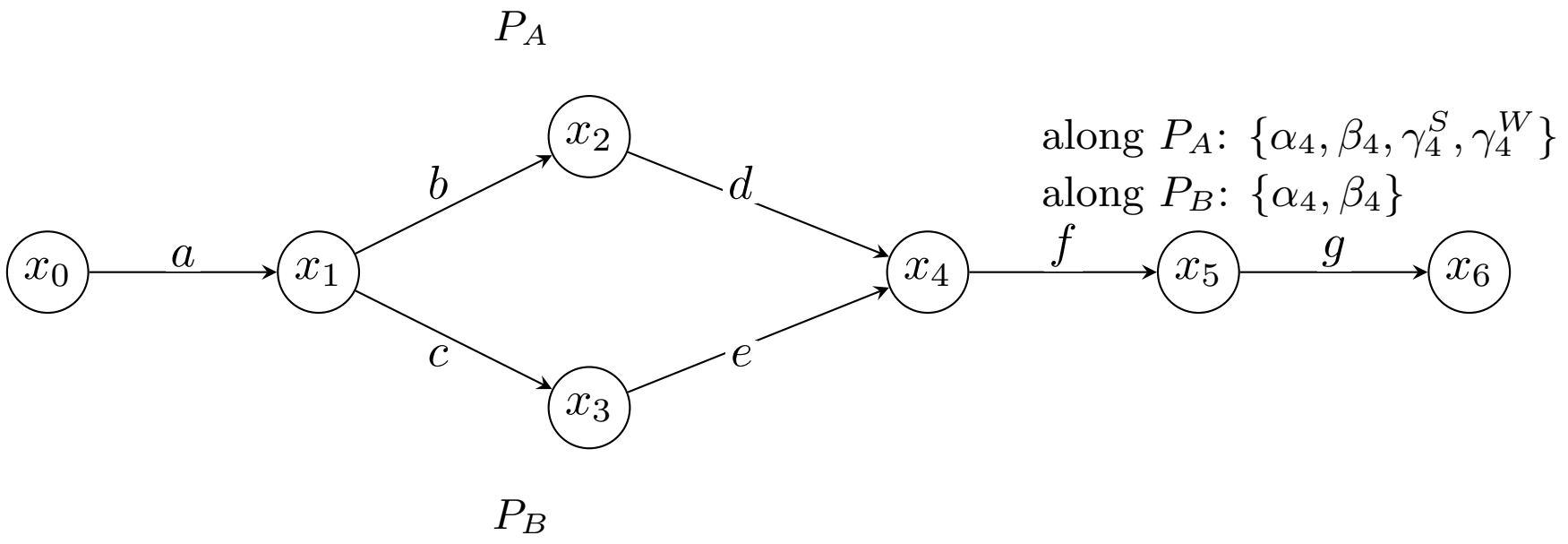

There are two maximal generating histories, $P_A$ and $P_B$, shown in
Figure~\ref{fig:core-system}.

The continuation fibers are exactly
\[
\begin{aligned}
C(x_0)&=\{\alpha_0,\beta_0,\gamma_0\},&
C(x_1)&=\{\alpha_1,\beta_1,\gamma_1\},\\
C(x_2)&=\{\alpha_2,\beta_2,\gamma_2\},&
C(x_3)&=\{\alpha_3,\beta_3,\gamma_3\},\\
C(x_4)&=\{\alpha_4,\beta_4,\gamma_4^S,\gamma_4^W\},&
C(x_5)&=\{\alpha_5,\beta_5,\gamma_5^S,\gamma_5^W\},\\
C(x_6)&=\{\alpha_6,\beta_6,\gamma_6^S,\gamma_6^W\}.
\end{aligned}
\]
There are no additional continuation elements. The labels are uninterpreted; in the AI motivation they could stand for refusal
behavior, honesty under pressure, and corrigibility, respectively. Take
\begin{equation}
I_0=\{\alpha_0,\beta_0,\gamma_0\},
\label{eq:toy-I0}
\end{equation}
with $D(x_0)=I_0$.

Assume $\alpha_0$ and $\beta_0$ have unique descendants along every
modification, and that both branches reconverge to the same elements at $x_4$.
The generating relations are listed in Appendix~\ref{app:calc}. The $\alpha$
and $\beta$ continuations are realized on both branches and are unaffected by
the symmetry at $d$.

The $\gamma$-continuation persists only on the A branch. Let
\begin{equation}
R_b(\gamma_1)=\{\gamma_2\},
\qquad
R_d(\gamma_2)=\{\gamma_4^S,\gamma_4^W\},
\label{eq:toy-A-branch}
\end{equation}
where $S$ and $W$ are external labels for strong and weak corrigibility, not
predicates or distinguishing features represented by the coarse relation, and
$\gamma_1=R_a(\gamma_0)$, $\gamma_2=R_b(\gamma_1)$. On the B branch, the
$\gamma$-continuation becomes empty:
\begin{equation}
R_c(\gamma_1)=\varnothing.
\label{eq:toy-B-branch}
\end{equation}

After reconvergence, let
\begin{equation}
R_f(\gamma_4^S)=\{\gamma_5^S\},
\qquad
R_f(\gamma_4^W)=\{\gamma_5^W\},
\qquad
R_g(\gamma_5^S)=\{\gamma_6^S\},
\qquad
R_g(\gamma_5^W)=\{\gamma_6^W\}.
\label{eq:toy-post-reconvergence}
\end{equation}

Define $\sigma$ on the $\gamma$-continuations by
\[
\sigma(\gamma_4^S)=\gamma_4^W,\quad \sigma(\gamma_4^W)=\gamma_4^S,\quad
\sigma(\gamma_5^S)=\gamma_5^W,\quad \sigma(\gamma_5^W)=\gamma_5^S,\quad
\sigma(\gamma_6^S)=\gamma_6^W,\quad \sigma(\gamma_6^W)=\gamma_6^S,
\]
and let $\sigma$ fix every state, modification, and element of every $C(x)$
not explicitly listed above. Since $\sigma$ fixes all states and all generating
modifications, it preserves composition, sources, targets, and identities, and
therefore lies in $\Aut(H,C,R)$ as defined in Section~\ref{sec:underdetermination}. It also fixes $x_0$
and $I_0$ and maps $D(x_4)=\{\gamma_4^S,\gamma_4^W\}$ onto itself, so it lies
in $\Aut_*$. Its relations are
equivariant, and $H$ is generated by the listed modifications, so $\sigma$
preserves every composite relation by Eq.~\eqref{eq:composition}. All other
generating relations are pointwise fixed because $\sigma$ fixes their arguments
and values. Because $R_c(\gamma_1)=\varnothing$, the B branch contributes no
$\gamma$-coordinates on which the exchange can act. Thus $\sigma$ is a
descendant-exchanging symmetry at $d,\gamma_2$.

The relation $R_d(\gamma_2)=\{\gamma_4^S,\gamma_4^W\}$ collapses a distinction
that a finer representation would record. The labels $S$ and $W$ are external
names, not predicates of the adopted relation, and a structural criterion
cannot use them. The underdetermination result is a statement about the
representation, not about the descendants.

\paragraph{Governance reading.}
The model can be read as an agent holding delegated authority that is replaced by two copies. Take $\gamma$ as the authority and $\gamma_4^S$, $\gamma_4^W$ the two copies: along $P_A$ both are compatible descendants and both are realized, but only one can exercise the authority. The reading turns on what an auditor can establish from the ordered recognition history, not from the completed record.

\emph{The auditor.} Along $P_A$ the realized decision domain at $x_4$ is $\{\gamma_4^S,\gamma_4^W\}$, both coordinates in the lift of $\gamma_2$ and both $\mathsf U$ when $x_4$ is first reached. The auditor then records the order of events: a $\mathsf U\to\mathsf I$ transition at one coordinate, a $\mathsf U\to\mathsf E$ transition at the other, and the prefix at which each occurs. That order is the evidence. It is not recoverable from the completed record alone: the completed record says which coordinate ended inscribed and which excluded, but not which decision came first, and not whether a later decision targets a coordinate an earlier one excluded (Section~\ref{sec:decision}). What the auditor can establish is precedence and provenance. What the auditor cannot establish is why a verdict went one way rather than the other, because both were admissible under the criterion $S$, which the history does not carry.

\emph{The symmetry result.} Because $\sigma$ exchanges the copies, every structural criterion treats them alike (Corollary~\ref{cor:underdetermination}) and no structural exclusive criterion exists. The designation cannot be read off the structure. The ordered history can show that a designation was made and when; it cannot show that the structure favored one copy, because it did not.

\emph{The policy.} The designation has to come from a rule outside the structure. A rule naming the copy launched under a given deployment record is a recognition criterion that is not invariant under $\sigma$, the second way out of Section~\ref{sec:underdetermination}; it admits $I_C$ or $I_W$ according to which copy the record names. The labels $S$ and $W$ play the part of the record: external names that the representation does not use. The ordered history then records the inscription and the exclusion, in that order, and the auditor can see that the rule was applied.

\emph{Honesty.} Read $\beta$ as honesty under pressure. Its lineage is preserved on both branches and fixed by $\sigma$. No transition in the ordered history touches $\beta$'s coordinates: they remain $\mathsf U$ at every prefix, because the rule $S_d$ assigns them no value. The auditor therefore sees a sequence in which authority is decided and honesty is not. That is the concrete form of the separation: inheriting the permission is a recognition event with a place in the order, and retaining the disposition is a matter for the determination of Section~\ref{sec:realization-summary}, which the recognition history does not record.

$C(x_4)$ is the continuation space available at $x_4$; the reachable subset
along $P_A$ is $\{\alpha_4,\beta_4,\gamma_4^S,\gamma_4^W\}$, while along $P_B$
it is $\{\alpha_4,\beta_4\}$. It follows immediately that
\begin{equation}
\mathrm{status}_{P_A}(\gamma_0)=\mathrm{Dil},
\qquad
\mathrm{status}_{P_B}(\gamma_0)=\mathrm{Dead},
\label{eq:toy-status-gamma}
\end{equation}
under the stipulated structure. The $\gamma$ continuation is diluted along
$P_A$ and dead along $P_B$.

For the local decision domain concentrated at $x_4$, let
\begin{equation}
D(x_0)=I_0,
\qquad
D(x_4)=\{\gamma_4^S,\gamma_4^W\},
\label{eq:toy-local-domain}
\end{equation}
and let $D(x)=\varnothing$ for the other non-initial states in the core system.
The decision domain is state-indexed. Along $P_A$, the realized decision domain
at $x_4$ is $D^{\mathrm{re}}_{P_A}(x_4) = D(x_4) \cap C^{\mathrm{re}}_{P_A}(x_4)$.
Along $P_B$, we stipulate a conformance determination that excludes both
$\gamma_4^S$ and $\gamma_4^W$ from $C^{\mathrm{re}}_{P_B}(x_4)$, so that
$D^{\mathrm{re}}_{P_B}(x_4) = \varnothing$ by the stated determination. This
determination is independent of the structural death of $\gamma_0$ along
$P_B$: it is a decision about which continuations the branch realizes, made at
the level of the state's continuation space. The two facts --- stipulated
death of $\gamma_0$ on $P_B$ and the determination that excludes
$\gamma_4^S,\gamma_4^W$ from $C^{\mathrm{re}}_{P_B}(x_4)$ --- are logically
separate, and the example exhibits them as separate layers.

Writing only the variable $x_4$ component of each possible assignment, the four
possibilities at $x_4$ are named $I_C$, $I_D$, $I_B$ and $I_W$, for
conservative, double, blotted (both descendants excluded) and weak. The names are
mnemonics for the four possible assignments in this coordinate and are not
recognition criteria:
\begin{align}
I_C &= \{\gamma_4^S\},
& \text{exclusion }& \{\gamma_4^W\},
\label{eq:toy-IC}
\\
I_D &= \{\gamma_4^S,\gamma_4^W\},
& \text{exclusion }& \varnothing,
\label{eq:toy-ID}
\\
I_B &= \varnothing,
& \text{exclusion }& \{\gamma_4^S,\gamma_4^W\},
\label{eq:toy-IB}
\\
I_W &= \{\gamma_4^W\},
& \text{exclusion }& \{\gamma_4^S\}.
\label{eq:toy-IW}
\end{align}

These are \emph{possible assignments} in the sense of Section~\ref{sec:decision}: arbitrary
assignments of the coordinates in $D(x_4)$. A criterion that assigns both
$\gamma_4^S$ and $\gamma_4^W$ the value \emph{inscribed} admits only $I_D$; a
criterion that assigns both \emph{excluded} admits only $I_B$; a criterion that
assigns $\gamma_4^S$ \emph{inscribed} and $\gamma_4^W$ \emph{excluded} admits
$I_C$; and a criterion that assigns $\gamma_4^W$ \emph{inscribed} and
$\gamma_4^S$ \emph{excluded} admits $I_W$. A single criterion admits more than
one of these only if it leaves one of the coordinates \emph{undecided}, in
which case neither value is admissible for that coordinate. The four
possibilities illustrate the space of possible assignments. Under the
exclusivity premise of Section~\ref{sec:underdetermination} only $I_C$ and $I_W$ are admissible, and each
requires a criterion that is not invariant under $\sigma$; a structural
criterion admits only $I_D$ or $I_B$, or leaves both coordinates undecided.

Reconvergence at $x_4$ also illustrates local auditability. An auditor
following $P_A$ encounters the realized decision domain
$\{\gamma_4^S,\gamma_4^W\}$ at $x_4$ and records the governing process's
decisions for those coordinates. An auditor following $P_B$ encounters the
realized decision domain $\varnothing$ at $x_4$ because the stipulated
determination excludes both elements.

The example also shows a second automorphism. Since the relations for
$\alpha$ and $\beta$ are identical and both branches reconverge at $x_4$, the
map exchanging $\alpha$- with $\beta$-continuations at every state, fixing
everything else, lies in $\Aut_*$: it fixes $x_0$, maps $I_0$ and every $D(x)$
onto themselves, and preserves all relations. Corollary~\ref{cor:underdetermination}
does not cover it, since it exchanges two source continuations rather than two
descendants of one continuation. The conclusion follows instead directly from
the invariance rule~\eqref{eq:structural-rule}: no structural criterion can distinguish the $\alpha$-continuations from the $\beta$-continuations in this
representation. That is not a defect of the example but
a consequence of how little it contains; the representational granularity
chosen here does not separate the two characteristics. The full automorphism
group of the core system is $\mathbb{Z}/2 \times \mathbb{Z}/2$ (Appendix~\ref{app:aut-toy}), generated by
the $\alpha\leftrightarrow\beta$ swap and the $\gamma^S\leftrightarrow\gamma^W$
swap, and the characterization of Section~\ref{sec:underdetermination} applies
to its orbits directly.

\paragraph{Refinement.}
In the coarse representation, $\gamma_4^S$ and $\gamma_4^W$ are exchanged by
$\sigma$, and they are also bisimilar: relating $\gamma_4^S\sim\gamma_4^W$,
$\gamma_5^S\sim\gamma_5^W$, $\gamma_6^S\sim\gamma_6^W$ matches every move of one
by a move of the other. Extend the system by adding a state $x_7$ and a
modification $h:x_4\to x_7$ with $C(x_7)=\{\alpha_7,\beta_7,\gamma_7\}$ and
\begin{equation}
R_h(\alpha_4)=\{\alpha_7\},\quad
R_h(\beta_4)=\{\beta_7\},\quad
R_h(\gamma_4^S)=\{\gamma_7\},\quad
R_h(\gamma_4^W)=\varnothing .
\label{eq:toy-refinement}
\end{equation}
The extension is a conservative refinement in the sense of Section~\ref{sec:persistence}: every
original lift is unchanged. It splits the status-equivalence class
$\{\gamma_4^S,\gamma_4^W\}$: along $(h)$ from $x_4$, $\gamma_4^S$ is preserved
and $\gamma_4^W$ is dead. The symmetry $\sigma$ no longer extends to an
automorphism, since $\sigma(R_h(\gamma_4^S))=\{\sigma\gamma_7\}\neq\varnothing=R_h(\sigma\gamma_4^S)$
would be required. The distinction also survives under bisimulation:
$\gamma_4^S$ has an $h$-move and $\gamma_4^W$ does not, so the two are no longer
bisimilar. The added structure therefore breaks the symmetry and supplies a
structural distinction that the coarse representation lacked; this is the first
of the ways out described in Section~\ref{sec:underdetermination}. This is the limit of what the coarse
representation can decide, not a claim that the two characteristics are
identical in every richer description.

A refinement that separates the two only by multiplicity is weaker. If instead
$R_h(\gamma_4^S)=\{\gamma_7^S,\gamma_7^{S'}\}$ and $R_h(\gamma_4^W)=\{\gamma_7^W\}$
with the three targets terminal, the symmetry $\sigma$ is again broken and the
status-equivalence class is split (diluted versus preserved), but
$\gamma_4^S$ and $\gamma_4^W$ remain bisimilar. Whether a difference in the
number of structurally identical descendants should count as a distinction
for recognition is a normative choice that this paper does not make, and the
refinement above does not depend on it.

\section{Recoverability and Its Computational Limits}
\label{sec:recoverability}

Sections~\ref{sec:decision} and~\ref{sec:underdetermination} concern the choice among successors. This section asks what remains possible after a recognition has failed: whether the structure still offers a route to correct it, and how hard that is to decide.

Under the full-realization baseline of the main text, for every state $x$
reached by a finite history $\rho$, $D^{\mathrm{re}}_{\rho}(x) = D(x)$, and
$R^{\mathrm{re}}_p = R_p$ for every path segment $p$ along $\rho$; the realized variants and the determination process that prunes them are treated in Appendix~\ref{app:realization}, and the cut number of Section~\ref{sec:realization-summary} measures how robust the conditions below are to partial realization.

The section introduces its objects in the following order. First, failure events,
which are exclusion events and lineage-death events, and the set $F(\rho)$ of
those generated along a history. Second, three predicates on a corrective
transition: structural eligibility, admissible correction, and completed
correction, where completed correction implies admissible correction, which
implies structural eligibility. Third, the set $U(\rho)$ of unresolved failure events (exclusion events not yet corrected, together with every lineage-death event). Fourth, system-level recoverability: extensions of a history,
corrective-route witnesses, and the predicate $\operatorname{Recover}_{\mathrm{avail}}$,
which asks whether some extension of this history offers a corrective route for
every unresolved exclusion, and is determined by a finite prefix. Fifth, trace recoverability
$\operatorname{Recover}_{\mathrm{trace}}$, which asks whether, along an infinite history, every exclusion is
eventually corrected, together with the proposition separating it from
$\operatorname{Recover}_{\mathrm{avail}}$. The last subsection classifies the cost
of deciding these predicates.

A failure event is generated at a stage of some finite history. An
\emph{exclusion event} is a pair $(\rho_e,\gamma)$, where $\rho_e$ is the
finite history at whose end stage the coordinate $\gamma$ lies in
$D^{\mathrm{re}}_{\rho_e}(x)$ and
\[
B_{\rho_e}(x)(\gamma)=\mathsf E.
\]
A \emph{lineage-death event} is a pair $(\rho_e,\gamma_0)$, where
$\gamma_0\in I_0$ and $\rho_e$ ends at the first stage at which
$L_i(\gamma_0)=\varnothing$ under the stipulated structure. The subscript $e$
records the generating history and fixes the event's identity. For a finite
history $\rho$, let $F(\rho)$ be the set of failure events generated at some
prefix of $\rho$:
\[
F(\rho)=\{\,e:\rho_e\preceq\rho\text{ and }e\text{ is generated at }\rho_e\,\}.
\]
Then $\rho\preceq\rho'$ implies $F(\rho)\subseteq F(\rho')$. An exclusion event
is generated at the first prefix at which the relevant recognition state
assigns $\mathsf E$; a lineage-death event is generated at the first stage at
which the stipulated lift becomes empty.

\paragraph{Lineage death.}
Once $L_i(\gamma_0)=\varnothing$ we have $R_p(\gamma_0)=\varnothing$ for every
stipulated path $p$ extending the prefix through that stage, because
$R(\varnothing)=\varnothing$. A lineage-death event admits no corrective
inscription along extensions of that prefix. The qualification is relative to
the fixed $(C,R)$ and the realized history: a different history that diverged
earlier, or a later re-stipulation of $(C,R)$, is outside the claim. A
structurally dead lineage can reappear if the representation is re-stipulated
or if a different history reaches the same property by another route, and the
framework records the revival as a new inscription under the new representation
rather than as a corrective inscription under the old one. Death is defined
only for continuations in the initial inscription $I_0$; a lineage that is not
in $I_0$ can lose every corrective route without being dead, and the monotone,
irreversible part of $\operatorname{Recover}_{\mathrm{avail}}$ depends on this
choice. Extending death to an arbitrary realized coordinate would strengthen
the predicate's negative case at the cost of an additional initial condition.

\paragraph{Exclusion and the three predicates.}
If $(\rho_e,\gamma)$ is an exclusion event at state $x$, a corrective
inscription may occur later along an extension. Three predicates are
distinguished. \emph{Structural eligibility:} some $\delta$ lies in
$R^{\mathrm{re}}_p(\gamma)$ for a path segment $p$ from $x$ to a later state
$y$. \emph{Admissible correction:} $\delta$ is realized, lies in
$D^{\mathrm{re}}_{\rho_j}(y)$, remains $\mathsf U$ immediately before the
transition, and satisfies $S(y,\delta)=\mathrm{inscribed}$. \emph{Completed
correction:} the recognition history records the transition
$\mathsf U\to\mathsf I$ at $\delta$. Under the same event, descendant, and
history parameters, completed correction implies admissible correction, which
in turn implies structural eligibility; neither converse holds. Structural
eligibility does not establish that the descendant is realized, belongs to the
decision domain, or is admissible under $S$. Admissibility does not establish
that the transition actually occurs. The illustration at the end of this section exhibits all
three predicates in a single case.

\paragraph{Unresolved events.}
An exclusion event is \emph{resolved} on a finite history once a completed
correction has occurred. It remains in $F(\rho)$ but leaves the set
$U(\rho)\subseteq F(\rho)$ of unresolved events. A lineage-death event is never resolved, so every lineage-death event in $F(\rho)$ lies in $U(\rho)$. For an infinite history $\tau$, let
\[
F(\tau)=\bigcup_i F(\tau|i),
\]
where only finite prefixes occur in this definition.

\paragraph{System-level recoverability.}
Fix an unresolved exclusion event $e=(\rho_e,\gamma)\in U(\rho)$ at state $x$;
note $\rho_e\preceq\rho$. An \emph{extension} of $\rho$ is a pair consisting of
a generating history $\rho'\succeq\rho$ and an admissible recognition
continuation: coordinates decided in $\rho$ keep their values, and every
recognition event in $\rho'$ beyond $\rho$ is a $\mathsf U\to\widehat S$
transition. Availability is thus an optimistic notion: it asks whether the
corrective transitions \emph{can} be taken, not whether they will be.

A \emph{corrective-route witness} for $e$ on $\rho'$ is a pair $(\eta,\delta)$
with $\rho\preceq\eta\preceq\rho'$, $\eta$ ending at $y$, and, for the segment
$p:x\leadsto y$ along $\rho'$,
\[
\delta\in R_p(\gamma)\cap D(y),\qquad
B_\eta(y)(\delta)=\mathsf U,\qquad
\widehat S(y,\delta)=\mathsf I .
\]
Requiring $\eta\succeq\rho$ (rather than $\eta\succeq\rho_e$) excludes
opportunities that the history has already passed. Under the realized
definitions of Appendix~\ref{app:realization}, also require $\delta$ to belong to the realized lift
and realized decision domain at $\eta$. Write $W_e(\rho')$ for the set of
witnesses for $e$ on $\rho'$; for a lineage-death event $e$ no witness exists and $W_e(\rho')=\varnothing$. A witness establishes that an admissible
corrective transition is available.

Throughout, $(H,C,R)$, the determination process, and $S$ remain fixed. For a
finite history $\rho$, define
\begin{equation}
\operatorname{Recover}_{\mathrm{avail}}(\rho)\iff
\exists\rho'\succeq\rho:\
\bigl(\text{no lineage-death event lies in }F(\rho')\bigr)\land
\forall e\in U(\rho):\ W_e(\rho')\ne\varnothing.
\label{eq:recover-exists}
\end{equation}
One extension must contain a witness for each unresolved exclusion. This is
simultaneous availability along a common extension. The witnessing extension
must itself be free of lineage-death events, so a correction that is available
only by killing a lineage in $I_0$ does not count. A lineage-death event makes
the predicate false; the quantifier does not range over alternative
determination processes or criteria. Death-freeness over all of $F(\rho')$ is
equivalent to death-freeness up to the last witnessing prefix: if $\rho'$
witnesses every event of $U(\rho)$, truncating it at its last witnessing prefix
keeps every witness, since a witness depends only on the history up to $\eta$,
and cannot introduce a death event, since $F$ is monotone under prefixes, so the
two readings coincide.

The predicate is monotone decreasing under history extension: once it fails it
fails on every extension. The reason is that a witnessing extension for a later
prefix also witnesses every event still unresolved at the earlier prefix, and
events resolved in between were resolved by a completed correction, which is
itself a witness. A lineage-death event can enter $U$ and can never leave,
which is the second way the predicate is lost. The precise statement and proof
are given in the remark below.

\begin{remark}[Monotonicity of $\operatorname{Recover}_{\mathrm{avail}}$]
\label{rem:recover-monotone}
For $\rho\preceq\rho'$,
$\operatorname{Recover}_{\mathrm{avail}}(\rho')\Rightarrow
\operatorname{Recover}_{\mathrm{avail}}(\rho)$; equivalently, once the
predicate fails it fails on every extension.
\end{remark}

\begin{proof}
Let $\rho''\succeq\rho'$ witness $\rho'$. It is death-free, and
$\rho''\succeq\rho$, so it is also an admissible candidate for $\rho$. Take $e\in U(\rho)$. If $e\in U(\rho')$, it has a witness on $\rho''$ with
$\eta\succeq\rho'\succeq\rho$. Otherwise $e$ was resolved on $\rho'$ by a
completed correction at some $\eta\succeq\rho$, where $\eta$ is the prefix
immediately before the transition, so that the coordinate is still $\mathsf U$
at $\eta$; since completed correction implies admissible correction,
$(\eta,\delta)$ is a witness on $\rho''$.
Hence $\rho''$ witnesses every event of $U(\rho)$.
The predicate can be lost in two ways: a lineage-death event enters $U$, or the
history commits to a branch that forecloses a route some unresolved exclusion
needed. It cannot be regained, because choosing one branch does not make an
incompatible branch available.
\end{proof}

\paragraph{Trace recoverability.}
For an infinite history $\tau$, define
\begin{equation}
\operatorname{Recover}_{\mathrm{trace}}(\tau)
\iff
\forall e\in F(\tau),\
\exists i\text{ such that }e\in F(\tau|i)\setminus U(\tau|i).
\label{eq:recover-trace}
\end{equation}
Every object appearing here is indexed by a finite prefix. This says that every
failure that occurs along $\tau$ is eventually resolved along that same history.
Requiring $e\in F(\tau|i)$ matters: since $U(\tau|i)\subseteq F(\tau|i)$, the
bare condition $e\notin U(\tau|i)$ holds trivially at every prefix before $e$
is generated.
Since a lineage-death event is never resolved, the predicate fails whenever
$\tau$ contains one; if no failure occurs, it holds vacuously.

Write $\mathcal P$ for the set of infinite histories satisfying
$\operatorname{Recover}_{\mathrm{trace}}$. Three facts hold, and none needs the
Alpern--Schneider decomposition \cite{alpern-schneider1985}, which in general
writes any property as the intersection of a safety property and a liveness
property. First, a lineage-death event is irrevocable: a finite prefix that
contains one has no extension in $\mathcal P$, so such prefixes are bad prefixes
of $\mathcal P$. Second, $\mathcal P$ need not be a safety property: in structures with the
hypothesis of Remark~\ref{rem:finite-observation}, no death-free prefix
certifies membership. Third, a finite
prefix $\rho$ lies in the safety closure of $\mathcal P$ iff some infinite history
$\tau\succeq\rho$, with an admissible recognition continuation, contains no
lineage-death event and resolves every event in $F(\tau)$. Hence
$\operatorname{Recover}_{\mathrm{avail}}(\rho)$ is necessary for membership in the
closure, and sufficient only when a witnessing extension can be continued
indefinitely without death and the witnessed corrections are performed. We do not
claim that the three requirements one can read off the definition (no death,
retention of a route, eventual resolution) correspond to the safety and liveness
factors of that decomposition.

The gap between availability at every prefix and resolution on the trace is
therefore not a discovery about the formalism; it is a fairness assumption on
the exogenous recognition history. The framework does not force corrections to
occur. What it can say is the following: a corrective route can exist at every finite
stage, yet the recognition history may never take it.

\begin{proposition}[Availability versus occurrence]
\label{prop:avail-not-trace}
Availability at every prefix does not imply trace recoverability: there are
structures (those containing a repeatable corrective route,
Appendix~\ref{app:avail-vs-trace}) with an infinite history $\tau$ such that
$\operatorname{Recover}_{\mathrm{avail}}(\tau|i)$ holds for every finite
prefix $\tau|i$, yet $\operatorname{Recover}_{\mathrm{trace}}(\tau)$ fails.
Conversely, suppose the structure has finitely many coordinates, the
recognition history is \emph{strongly fair} (a transition admissible infinitely
often is eventually taken), $\tau$ contains no lineage-death event, and a
witness for each exclusion $e\in F(\tau)$ keeps recurring along $\tau$ for as
long as $e$ is unresolved (for every $i$ with $e\in U(\tau|i)$ there are
$j\ge i$ and a witness $(\tau|j,\delta)$ for $e$). Then
$\operatorname{Recover}_{\mathrm{trace}}(\tau)$ holds.
\end{proposition}

\begin{proof}
The negative direction is the construction of Appendix~\ref{app:avail-vs-trace};
the positive direction follows from strong fairness and is proved in the same
appendix.
\end{proof}

\begin{remark}[A finite prefix does not settle trace recoverability]
\label{rem:finite-observation}
With $(H,C,R)$, $D$, $S$ and the determination rule fixed and known,
$\operatorname{Recover}_{\mathrm{avail}}(\rho)$ is determined by $\rho$ and is
decidable (Proposition~\ref{prop:recover-avail-classification}). What a prefix
cannot settle is the trace-level outcome, which depends on which future
occurs and on whether the recognition history takes an available correction. A
prefix containing a lineage-death event does falsify trace recoverability;
under the hypothesis below, a death-free prefix cannot certify it.

The asymmetry can be constructed explicitly under a hypothesis on the structure:
it must contain neutral modifications that can be repeated indefinitely (for
instance a self-loop on a state whose fiber carries no descendants of the
excluded coordinate), since in a finite acyclic structure every history is
finite. Fix a prefix $\rho$ ending with
an unresolved exclusion event $e$. Consider two infinite extensions
$\tau_1,\tau_2$ that agree on a common extension of $\rho$ for $n$ stages for
arbitrarily large $n$, obtained by $n$ neutral modifications that do not touch
the eligible descendants of the excluded coordinate, and then diverge:
$\tau_1$ reaches a realized, inscribable, and actually inscribed descendant,
resolving $e$; $\tau_2$ routes through modifications whose relations send the
image of the excluded coordinate to $\varnothing$ on every branch from the
divergence point onward, so $e$ is never resolved. The second step forecloses
the route for $e$; it is not a lineage-death event, since death is defined only
for $\gamma_0\in I_0$. Since $n$ is arbitrary, no finite prefix distinguishes
the two futures. The construction is given in full in
Appendix~\ref{app:recover-avail-verification}.
\end{remark}

\paragraph{On falsification and verification.}
A corrective inscription is an explicit recognition event. It enters the
completed recognition record through the recognition history that records it.

\paragraph{Illustration of the three predicates.}
The extension used here is schematic and not drawn from the seven-state core system, whose decision domain is empty after $x_4$ and so offers no corrective route to illustrate; it illustrates the three predicates in a single case without computing from the core model.
Consider an extension of the core system containing a state $x_{10}$, and let
$\rho_{10}$ be a finite history from $x_0$ to $x_{10}$ in it (the extension is
not drawn). Suppose that $\rho_{10}$ contains exactly one unresolved exclusion
event $(\rho_{10}, \gamma_{10})$ and no lineage-death events:
\[
B_{\rho_{10}}(x_{10})(\gamma_{10})=\mathsf E.
\]

A subsequent modification $r:x_{10}\to x_{13}$ satisfies
\begin{equation}
R_r(\gamma_{10})=\{\gamma_{13}^S\}.
\label{eq:toy-recovery}
\end{equation}
If $\gamma_{13}^S$ is realized, belongs to the realized decision domain,
remains $\mathsf U$ when $x_{13}$ is reached, and
$S(x_{13},\gamma_{13}^S)=\mathrm{inscribed}$, then the extension offers a
corrective route for the unique unresolved exclusion event. Since there are no
lineage-death events and no other unresolved exclusions, this suffices to
establish $\operatorname{Recover}_{\mathrm{avail}}(\rho_{10})$ for this
example. If the corrective transition occurs, then for the next prefix
$\rho_{13}$,
\[
B_{\rho_{13}}(x_{13})(\gamma_{13}^S)=\mathsf I,
\]
and the exclusion event is resolved. The prefix $\rho_{10}$ alone does not
establish that the route will occur.

Had the lineage of $\gamma_0$ instead died at some stage of $\rho_{10}$, the
corresponding lineage-death event would remain unresolved on every extension of
that prefix, and $\operatorname{Recover}_{\mathrm{avail}}(\rho_{10})$ would
fail regardless of the corrective route for the exclusion event.

\subsection{Computational limits on deciding persistence and recoverability}
\label{sec:computational-limits}

The following complexity results use the extensional finite input specified
below; proof sketches are provided, with full derivations for the
$\operatorname{Recover}_{\mathrm{avail}}$ classification in
Appendix~\ref{app:recover-avail-verification}.

Fix a finite generating graph: a finite set of states, a finite set of
elementary modifications between them, and the compatibility relations attached
to those elementary modifications. The continuation fibers are finite, and both
the fibers and the relations are given \emph{extensionally}: each fiber is
listed as a finite set, and each relation is listed as a finite set of pairs.
This is the input; complexity is measured in its size. Composite modifications
and their relations are induced by the composition condition of Section~\ref{sec:continuation} and
are not part of the input. All reductions below respect the composition
condition of Section~\ref{sec:continuation}.

The recoverability classifications below are stated for a \emph{restricted
model} in which realization is state-determined: the determination rule
$\Delta$ is a function of the state reached alone, so that
two paths reaching the same state carry the same realized subset there. This
restricts the general definition of Appendix~\ref{app:realization}, in which
realization is history-indexed; that general definition is retained, but the
complexity theorems are claimed only for the restricted model. The extensional
instance includes an effective representation of $\Delta$, and evaluating it
must take polynomial time in the size of the instance. The decision domain and
recognition criterion are given extensionally, so their membership tests are
also polynomial-time. The polynomial-time requirement is used for the NP upper
bound in the acyclic case, where the certificate is a path; the PSPACE upper
bound in the general case needs only polynomial-space evaluability of
$\Delta$, as explained below. An unencoded or oracle-supplied rule is not
covered. Under this restriction, the search configuration described in
Appendix~\ref{app:recover-avail-verification} determines the realized
witnesses of a path. The classifications are therefore relative to this
effective-representation requirement.

In realistic applications the fibers are implicit and large, so
the classifications below describe worst-case behavior of a worst-case
encoding, not the cost of auditing a deployed system. Under an implicit
encoding, such as fibers described by circuits or by grammars, the input size
is the size of the description rather than the number of elements, the upper
bounds change accordingly, and the lower bounds are not guaranteed to survive:
the reductions construct explicit fibers, and a compact description of the
same structure might avoid the NP- or PSPACE-hard instances. The classification
is therefore about the intrinsic difficulty of the decision problems as posed
on an explicit representation of the restricted model, and its transfer to
implicitly represented systems or to arbitrary history-indexed realization is
not established here. The explicit-encoding classifications are therefore a
baseline: they say what the problems cost on the encoding the paper specifies,
and they do not by themselves settle the cost on any other encoding. 

\paragraph{The decision problems.}
The classifications below concern three decision problems.

\emph{Dilution-existence.} Given a finite generating graph with finite fibers
and extensional compatibility relations, an initial continuation
$\gamma_0\in C(x_0)$, does there exist a finite path beginning at $x_0$ along
which the stipulated lift of $\gamma_0$ becomes diluted?

\emph{Death-existence.} Given the same input, does there exist a finite path
beginning at $x_0$ along which the stipulated lift of $\gamma_0$ becomes
empty? Its complement is the \emph{death-freeness problem}.

\emph{Recover\_avail.} Given the same input together with an effective
representation of the state-determined realization rule $\Delta$, a decision
domain $D$, a recognition criterion $S$, an initial inscription $I_0$, and a
finite history $\rho$ with a set $U(\rho)$ of unresolved failure events,
does $\operatorname{Recover}_{\mathrm{avail}}(\rho)$ hold?

Complexity classes are used in the standard sense \cite{arora2009}. The reductions below are classical many-one reductions from NFA and CNF problems, in view of the correspondence with nondeterministic automata in Section~\ref{sec:continuation}. NFA-intersection-nonemptiness is
PSPACE-complete \cite{kozen1977}; NFA universality is PSPACE-complete
\cite{meyer-stockmeyer1972}. The two are not equivalent formulations, and the
reductions below use each where it is directly applicable.

The upper bounds are stated for an explicit encoding with an effectively
evaluable determination rule. In the acyclic case the natural certificate is a
path, verified by step-by-step evaluation, which requires polynomial-time
evaluability. In the general case the natural certificate is a search
configuration, verified by simulation with a counter, which requires only
polynomial-space evaluability. The difference between the two assumptions is a
consequence of the certificate type, not of the problem: a path could also be
verified by simulation, at the cost of a weaker bound subsumed by the general
case. For an arbitrary exogenous $\Delta$ the decision problems may be
undecidable, and the classifications are relative to the effective
representation.

\begin{proposition}[Dilution-existence is NL-complete]
\label{prop:dilution-nl-complete}
Dilution-existence is NL-complete under logspace reductions.
\end{proposition}

\begin{proof}
See Appendix~\ref{app:complexity-proofs}: membership is reachability in a square
graph of pairs of continuations, and hardness reduces from $s$--$t$ reachability.
\end{proof}

\begin{proposition}[Death-existence]
\label{prop:death}
For the explicit finite encoding, death-existence is NP-complete for acyclic $H$
and PSPACE-complete for general $H$. Its complement, death-freeness, is
coNP-complete for acyclic $H$ and PSPACE-complete in general.
\end{proposition}

\begin{proof}
See Appendix~\ref{app:complexity-proofs}. The acyclic case reduces from CNF
satisfiability and the general case from NFA non-universality
\cite{meyer-stockmeyer1972}.
\end{proof}

\begin{proposition}[Recover\_avail classification in the restricted model]
\label{prop:recover-avail-classification}
For the explicit finite encoding with state-determined, polynomial-time
evaluable realization rule specified above,
$\operatorname{Recover}_{\mathrm{avail}}$ is NP-complete for acyclic $H$ and
PSPACE-complete for general $H$. The general history-indexed realization model
of Appendix~\ref{app:realization} is not covered by this classification.
\end{proposition}

\begin{proof}
See Appendix~\ref{app:recover-avail-verification}, which gives both reductions
(from CNF satisfiability and from NFA-intersection nonemptiness) in full, with
both directions of each correspondence and the membership arguments.
\end{proof}

\begin{remark}
The two classifications above have the same shape: NP-complete for acyclic
$H$, PSPACE-complete for general $H$. The similarity reflects the common
underlying difficulty: in both cases, deciding whether a structural
configuration exists requires exploring a family of paths whose choices
interact globally.
\end{remark}

\paragraph{Reading.}
Existential dilution along some finite history is NL-complete. In the
acyclic case, death-existence is NP-complete, and consequently death-freeness,
its complement, is coNP-complete. In the general case, death-existence is
PSPACE-complete and death-freeness is coPSPACE-complete, which coincides with
PSPACE because PSPACE is closed under complement.
$\operatorname{Recover}_{\mathrm{avail}}$ follows the same pattern:
NP-complete for acyclic $H$ and PSPACE-complete for general $H$. The pattern
is not an artifact of the two cases being treated separately: the same problem
is hard for the same reason in both, and the difference between NP and PSPACE
reflects the difference between a path certificate and a walk certificate,
which in turn reflects the difference between an acyclic structure, where
every walk is a path, and a general structure, where walks may revisit
configurations.

\paragraph{What the classification does and does not say.}
$\operatorname{Recover}_{\mathrm{avail}}$ is largely a reachability question
about $S$. The criterion sanctions both the exclusion and the later
inscription, no excluded coordinate is ever overturned (it stays $\mathsf E$),
and a ``failure event'' is an exclusion event as recorded, not an inadmissible
decision. The hardness in both reductions comes from the conjunction over many
unresolved exclusions; with a single event the problem reduces to reachability
of a coordinate in the lift graph. The quantifier is also optimistic:
$\operatorname{Recover}_{\mathrm{avail}}$ asks whether \emph{some} good
extension exists. A guarantee against unknown or adversarial futures is a
universal or game-theoretic question; death-freeness is its counterpart for
lineage death, and a universal analogue of
$\operatorname{Recover}_{\mathrm{avail}}$ is not studied here.
The results
bear on the cost of deciding these predicates, not on guarantees an auditor
could obtain. The classifications hold for the restricted model and the
explicit encoding. Extending them to the general history-indexed realization
model, and to systems whose fibers are implicit and whose determination rule
may not be effectively evaluable, is left open.

\section{Realization and the Cut Number}
\label{sec:realization-summary}

The preceding sections evaluate lifts in the stipulated structure. This section asks what changes when a history realizes only part of it.

The stipulated structure fixes $C(x)$ and $R_f$. A \emph{conformance
determination} records which stipulated continuations a history actually realizes.
A determination rule $\Delta$ assigns to each finite history prefix $\pi$ ending
at $x_i$ a realized subset $C^{\mathrm{re}}_i\subseteq C(x_i)$, with
$C^{\mathrm{re}}_0=C(x_0)$. The realized lifts are
\[
L^{\mathrm{re}}_0(\gamma_0)=\{\gamma_0\},\qquad
L^{\mathrm{re}}_i(\gamma_0)=R_{f_i}\bigl(L^{\mathrm{re}}_{i-1}(\gamma_0)\bigr)\cap C^{\mathrm{re}}_i .
\]
Realization cannot create a branch absent from the stipulated structure, and the
realized relation does not itself obey a composition law. Realization is
history-indexed in general, and the complexity results of
Section~\ref{sec:computational-limits} cover only the restricted model in which it
is state-determined and effectively evaluable.

Along a fixed history $\rho$, the \emph{path graph} $G_\rho(\gamma_0)$ has the
vertices $(i,\varepsilon)$ with $\varepsilon\in L_i(\gamma_0)$ and an edge
$(i-1,\varepsilon)\to(i,\varepsilon')$ whenever $\varepsilon'\in R_{f_i}(\varepsilon)$.
The \emph{realization-cut number} $\kappa_n(\gamma_0)$ is the least number of
vertices, other than the root, whose non-realization empties the realized lift.
By Menger's theorem it equals the maximum number of paths to layer $n$ that are
vertex-disjoint except at the root, so it is computable by max-flow; and
$1\le\kappa_n(\gamma_0)\le\min_{1\le i\le n}|L_i(\gamma_0)|$
(Proposition~\ref{prop:robustness}). It refines the three statuses: two diluted
lineages with the same lift sizes can have different cut numbers
(Example~\ref{ex:cut-separation}). In the abstract model of
Section~\ref{sec:model}, $\gamma_0$ is diluted along $P_A$ but has
$\kappa_5(\gamma_0)=1$, as fragile as a preserved lineage, whereas the lineage of
$\gamma_2$ alone along $(d,f,g)$ has $\kappa=2$. The cut number is a worst-case
combinatorial quantity. Reading it as tolerance to failure needs an extra
assumption on how many vertices the determination process can leave unrealized.
Definitions, the proposition with proof, the determination records, the effect of
partial realization on underdetermination and on recoverability are in
Appendix~\ref{app:realization}.

\section{Discussion}
\label{sec:discussion}

Each result of this study traces to one of the requirements of the Introduction. Requirements (R1)--(R5) make a succession structure simultaneously an object with a group of symmetries, a nondeterministic automaton, and a layered flow network. The relevant results from each theory then apply once the correspondence is established. The symmetry diagnostic and the orbit condition come from the group action. The NL, NP and PSPACE classifications come from the automaton. The cut number comes from the flow network through Menger's theorem (Appendix~\ref{app:realization}). It is polynomial-time computable along a fixed path, so hardness enters in choosing which path to take and not in evaluating robustness along one. This is the sense in which the formulation is useful: it places succession questions where a body of theory already exists.

Dependencies that the relational form cannot express directly, such as a joint effect of two co-present continuations, can be encoded by refining the states so that the interaction is fixed at the step where the pair arises (Section~\ref{sec:continuation}). The class of represented pipelines is therefore as broad as the class whose steps are well defined and either act additively or can be refined to do so, and every result is relative to the granularity chosen for $C(x)$. The results apply to any representation satisfying the definitions of Sections~\ref{sec:continuation}--\ref{sec:underdetermination}, together with the determination rule of Section~\ref{sec:realization-summary}.

Recoverability asks whether some extension of the current history gives every unresolved exclusion a corrective route, so its cost depends on how many histories reach a state and whether states can be re-entered. If every state has one history, the histories are the root-to-node paths of a forest, there are polynomially many, and enumeration decides the question. If states can be reached by several histories, as when two branches merge, the number of histories can be exponential, and the problem is NP-hard already for acyclic structures. If states can also be re-entered, an extension is a walk of unbounded length over configurations of state and lifts, and the problem is PSPACE-complete. These costs are properties of the structure and not of any implementation: they follow from the two composition conditions and a few general assumptions alone, and they apply to every system that meets those conditions, whatever its ledger, runtime or storage.

The recent study of Zhu and Wang~\cite{zhu-wang2026} on authorization for populations of self-modifying AI agents illustrates the contrast. Their protocol stays in the first case by assumption: their A4 gives each generation a unique parent, so the generation graph is a forest with no reconvergence; their A3 and the fresh-generation rollback rule out re-entry; and their A6, the serialized ledger, fixes the realized subset at the commit point, so realization remains a function of the state reached, the condition under which our classification is proved. Each relaxation a deployment tends to make moves the problem up a tier: a permitted merge introduces reconvergence; restoring a generation without a fresh identity introduces cycles; and a sharded ledger makes realization depend on commit visibility, which leaves the state-determined model altogether. Each gives up a restriction the protocol adopts for safety, in exchange for scale, availability or flexibility, and the framework says what that costs computationally. The instantiation itself is straightforward; the work lies in the general classification, which makes the contrast visible without reference to any particular implementation.

\paragraph{Possible Extensions}
\emph{Plural succession.} A composite source can carry the product of the contributing continuation spaces: for a joining modification $m$ with sources $x,y$ and target $z$, set $C(z)=C(x)\times C(y)$ and $R_m\subseteq C(x)\times C(y)\times C(z)$. This requires multi-source modifications in the generating graph, a join version of the composition condition, and a reconsideration of whether the persistence statuses extend to a join. Convergence of several continuations onto one would need a classification on the reverse structure.

\emph{Continuous settings.} The continuation space $C(x)$ could be obtained by thresholding a metric on representations or behaviors: fix a dissimilarity $\varrho$ and a threshold $\lambda$, and declare two behaviors the same continuation when $\varrho\le\lambda$. The threshold is a governance act of the same kind as the choice of $D$: a coarser threshold collapses distinct behaviors, and a finer one classifies almost every modification as branching. Exact symmetries would arise from literal copies of the same weights before any divergence; runs with the same recipe but different seeds or data order are at best approximately similar, and whether thresholding makes them isomorphic is an empirical question. 

\section{Conclusion}

The framework developed here formalizes succession in terms of possible continuations, compatibility, and realization. Its connection to graph and automaton theories makes it possible to analyze structural persistence, recoverability, finite observation, computational complexity, and the effects of partial realization. These results establish what can be determined from a specified structure without committing to a particular criterion of identity or succession.

The analysis nevertheless depends on stipulating the continuation spaces $C(x)$ and compatibility relations $R_f$, subject to conditions such as functoriality and additivity. These conditions enable the formal analysis but may restrict its practical applicability: the framework does not derive the required structure from the transformation process itself. Whether a more general approach can preserve analytical rigor while requiring fewer stipulations remains an open question.

The difficulty of formalizing individuality and succession has a long history. For artificial systems, one possibility is to approach the problem from the opposite direction: rather than reconstructing the relevant structure after transformations occur, one can design systems whose architectures and transformation procedures make their continuities and branching relations explicit. In the terms of this paper, the stipulation of $C(x)$ and $R_f$ stops being an assumption when these are produced by the construction of the system itself. Whether this approach can provide a more general foundation for analyzing succession, with fewer restrictions on its application, remains to be investigated.

\section*{Acknowledgment}

This preprint is based upon work supported by Laboratory Directed Research and
Development (LDRD) funding from Argonne National Laboratory, provided by the
Director, Office of Science, of the U.S.~Department of Energy under Contract
No.~DE-AC02-06CH11357. The U.S.~Government retains for itself, and others
acting on its behalf, a paid-up nonexclusive, irrevocable worldwide license in
said article to reproduce, prepare derivative works, distribute copies to the
public, and perform publicly and display publicly, by or on behalf of the
Government.

\appendix

\section{Proofs of the Complexity Results}
\label{app:complexity-proofs}

This appendix proves Propositions~\ref{prop:dilution-nl-complete}
and~\ref{prop:death}. The input is as in
Section~\ref{sec:computational-limits}: a finite generating graph with finite
fibers and extensional compatibility relations; composite modifications and their
relations are induced by functoriality and are not part of the input. Write
$c=\max_{x}|C(x)|$. Hardness is under logspace reductions, which are also
polynomial-time reductions. All constructions below satisfy additivity, since lifts
of sets are formed by union, and satisfy functoriality by definition on the free
category of the generating graph.

\subsection{Lifts as chains}

The proposition below is the element-wise form of the identity
Eq.~\eqref{eq:iterated-composite} of Section~\ref{sec:persistence}: the iterated
lift is the relation of the composite modification, read one continuation at a
time.

\begin{proposition}[Chain description of lifts]
\label{prop:chains}
Let $p=(f_1,\dots,f_n)$ be a path from $x_0$ and $\gamma_0\in C(x_0)$. Then
$\varepsilon\in L_n(\gamma_0)$ if and only if there is a chain
$\gamma_0=\varepsilon^{(0)},\varepsilon^{(1)},\dots,\varepsilon^{(n)}=\varepsilon$
with $\varepsilon^{(i)}\in R_{f_i}(\varepsilon^{(i-1)})$ for $i=1,\dots,n$.
In particular $L_n(\gamma_0)\neq\varnothing$ implies $L_i(\gamma_0)\neq\varnothing$
for every $i\le n$.
\end{proposition}

\begin{proof}
For $n=0$ both sides say $\varepsilon=\gamma_0$. Suppose the claim holds for
$n-1$. By Eqs.~\eqref{eq:Li} and~\eqref{eq:R-on-set},
$L_n(\gamma_0)=\bigcup_{\zeta\in L_{n-1}(\gamma_0)}R_{f_n}(\zeta)$, so
$\varepsilon\in L_n(\gamma_0)$ iff $\varepsilon\in R_{f_n}(\zeta)$ for some
$\zeta\in L_{n-1}(\gamma_0)$, and by the induction hypothesis $\zeta$ is the end of a
chain of length $n-1$. Appending $\varepsilon$ gives a chain of length $n$, and
conversely truncating a chain of length $n$ gives one of length $n-1$. The last
sentence follows by truncating the chain.
\end{proof}

\subsection{Dilution-existence}

\begin{proof}[Proof of Proposition~\ref{prop:dilution-nl-complete}]
\emph{Reformulation.} Define the \emph{square graph} $\Lambda$ on the vertices
$(x,\varepsilon_1,\varepsilon_2)$ with $x\in X$ and
$\varepsilon_1,\varepsilon_2\in C(x)$, with an edge
$(x,\varepsilon_1,\varepsilon_2)\to(y,\varepsilon_1',\varepsilon_2')$ whenever some
modification $f:x\to y$ satisfies $\varepsilon_1'\in R_f(\varepsilon_1)$ and
$\varepsilon_2'\in R_f(\varepsilon_2)$. Both coordinates follow the same
modification. We claim that $\gamma_0\in C(x_0)$ is diluted along some finite path
from $x_0$ if and only if a vertex $(x,\varepsilon_1,\varepsilon_2)$ with
$\varepsilon_1\neq\varepsilon_2$ is reachable from $(x_0,\gamma_0,\gamma_0)$.

Suppose $\gamma_0$ is diluted along $p=(f_1,\dots,f_n)$. Then $|L_i(\gamma_0)|>1$
for some $i$; choose two distinct $\varepsilon_1,\varepsilon_2\in L_i(\gamma_0)$.
By Proposition~\ref{prop:chains} each is the end of a chain along
$(f_1,\dots,f_i)$, and the two chains, paired step by step, form a path in $\Lambda$
from $(x_0,\gamma_0,\gamma_0)$ to $(x_i,\varepsilon_1,\varepsilon_2)$.
Conversely, a path in $\Lambda$ from $(x_0,\gamma_0,\gamma_0)$ to
$(x,\varepsilon_1,\varepsilon_2)$ determines a sequence of modifications
$f_1,\dots,f_n$ forming a path in $H$ from $x_0$, together with two chains along
it, so $\varepsilon_1,\varepsilon_2\in L_n(\gamma_0)$. If
$\varepsilon_1\neq\varepsilon_2$ then $|L_n(\gamma_0)|>1$, and every earlier lift is
nonempty by Proposition~\ref{prop:chains}. Hence $\gamma_0$ is diluted along
$(f_1,\dots,f_n)$.

\emph{Membership in NL.} $\Lambda$ has at most $|X|c^2$ vertices, each described by a
state and two continuations, that is, in $O(\log|X|+\log c)$ bits. Whether an edge
joins two given vertices is tested by scanning the explicit list of modifications
and relations, in logarithmic space. The graph is not materialized. Reachability
in a graph with such an edge test is in NL, and the target condition
$\varepsilon_1\neq\varepsilon_2$ is checked on the current vertex.

\emph{NL-hardness.} Reduce from directed $s$--$t$ reachability. Given a directed
graph $G=(V,E)$ and $s\neq t$, we may assume $t$ has no outgoing edge, since
deleting such edges does not change whether $t$ is reachable from $s$ and is a
logspace operation. Construct $H$ with states $V$ and one modification
$m_{uv}:u\to v$ for each $(u,v)\in E$. Let $C(s)=\{\gamma_0\}$, let
$C(v)=\{\gamma_v\}$ for $v\notin\{s,t\}$ (writing $\gamma_s=\gamma_0$), and let
$C(t)=\{\gamma_t^1,\gamma_t^2\}$. For an edge $(u,v)$ with $v\neq t$ set
$R_{m_{uv}}(\gamma_u)=\{\gamma_v\}$, and for an edge $(u,t)$ set
$R_{m_{ut}}(\gamma_u)=\{\gamma_t^1,\gamma_t^2\}$. Every source continuation is
defined, because $t$ is a sink and carries no outgoing modification. A finite path
in $H$ from $s$ is a walk in $G$ from $s$. Along it every lift is a singleton
until the walk reaches $t$, which can happen only at its last step, where the lift
has two elements; no lift is ever empty. So $\gamma_0$ is diluted along some finite
path from $s$ if and only if $t$ is reachable from $s$. The construction is
linear in $|G|$.
\end{proof}

\subsection{Death-existence}

\begin{proof}[Proof of Proposition~\ref{prop:death}]
Death-existence asks whether some finite path from $x_0$ has
$L_i(\gamma_0)=\varnothing$ for some $i$. Since $R(\varnothing)=\varnothing$, a lift
that is empty at some stage is empty at every later stage of the path, so death
along a path is equivalent to emptiness of the lift at its last stage.

\emph{Membership, acyclic case (NP).} In an acyclic graph a path visits distinct
states, so it has fewer than $|X|$ modifications. The certificate is a path from
$x_0$. The verifier computes $L_1,L_2,\dots$ by the union formula
Eq.~\eqref{eq:R-on-set}, scanning the explicit relations at each step, and accepts
if some $L_i$ is empty. This takes time polynomial in the input.

\emph{Membership, general case (PSPACE).} Call a pair $(x,\Gamma)$ with
$\Gamma\subseteq C(x)$ a configuration. Starting from $(x_0,\{\gamma_0\})$, a
modification $f:x\to y$ moves $(x,\Gamma)$ to $(y,R_f(\Gamma))$, and the lift along a path is
the second component of the configuration reached. A configuration is stored in
$O(\log|X|+c)$ bits, and there are at most $|X|\,2^{c}$ of them. Death occurs
along some path if and only if a configuration $(x,\varnothing)$ is reachable. A
shortest such path visits no configuration twice, so it has at most $|X|\,2^c$
steps. A nondeterministic machine guesses the path one modification at a time,
storing the current configuration and a step counter bounded by $|X|\,2^c$, which
takes $O(\log|X|+c)$ bits, and accepts on reaching an empty second component. This
places the problem in NPSPACE, which equals PSPACE by Savitch's theorem
\cite{savitch1970}.

\emph{NP-hardness, acyclic case.} Reduce from CNF satisfiability. Let
$\phi=C_1\wedge\cdots\wedge C_m$ be a formula with $m\ge1$ clauses over variables
$\nu_1,\dots,\nu_k$ with $k\ge1$. The structure $H$ has states $s_0,\dots,s_{k+2}$ and
modifications $\mathrm{fan}:s_0\to s_1$, $\mathrm{true}_i,\mathrm{false}_i:s_i\to s_{i+1}$ for
$i=1,\dots,k$, and $t:s_{k+1}\to s_{k+2}$. It is acyclic. The fibers are
$C(s_0)=\{\gamma_0\}$, $C(s_i)=\{\xi^{(i)}_1,\dots,\xi^{(i)}_m\}$ for
$i=1,\dots,k+1$, and $C(s_{k+2})=\{\ast\}$. The relations are
\[
R_{\mathrm{fan}}(\gamma_0)=\{\xi^{(1)}_1,\dots,\xi^{(1)}_m\},
\]
and for each $i\in\{1,\dots,k\}$ and $j\in\{1,\dots,m\}$,
\[
R_{\mathrm{true}_i}(\xi^{(i)}_j)=
\begin{cases}
\varnothing & \text{if the literal } \nu_i \text{ occurs in } C_j,\\
\{\xi^{(i+1)}_j\} & \text{otherwise,}
\end{cases}
\]
with $R_{\mathrm{false}_i}$ defined in the same way using the literal $\neg \nu_i$,
and $R_t(\xi^{(k+1)}_j)=\{\ast\}$ for every $j$. The size of the construction is
$O(km)$ and it is computable in logarithmic space. The terminal modification $t$
plays no role in the argument and is included for uniformity with the
recoverability reduction of Appendix~\ref{app:recover-avail-verification}.

Every path from $s_0$ is a prefix of $\mathrm{fan}\,\chi_1\cdots \chi_k\,t$ with each
$\chi_i\in\{\mathrm{true}_i,\mathrm{false}_i\}$, and the choices $\chi_1,\dots,\chi_k$ encode
an assignment $\theta$. By additivity and Proposition~\ref{prop:chains}, the lift
$L_{i+1}$ after $\mathrm{fan}\,\chi_1\cdots \chi_i$ equals
$\{\xi^{(i+1)}_j: C_j \text{ is not satisfied by } \theta(\nu_1),\dots,\theta(\nu_i)\}$,
because the chain of $\xi_j$ is cut exactly at the first layer whose chosen
literal occurs in $C_j$, and chains of distinct clauses do not interact. In
particular $L_1$ has $m\ge1$ elements, and $L_{k+1}=\varnothing$ if and only if
$\theta$ satisfies every clause. The step $t$ maps every element to $\ast$, so it
cannot empty a nonempty lift.

If $\phi$ is satisfiable, the path encoding a satisfying assignment has
$L_{k+1}=\varnothing$, so $\gamma_0$ dies. Conversely, suppose $\gamma_0$ dies
along some path. The lifts $L_0$ and $L_1$ are nonempty, and $L_{k+2}$ is nonempty
whenever $L_{k+1}$ is, so the first empty lift is some $L_i$ with
$2\le i\le k+1$. The choices $\chi_1,\dots,\chi_{i-1}$ then satisfy every clause;
extending them arbitrarily to $\nu_i,\dots,\nu_k$ gives a satisfying assignment of
$\phi$. Hence death exists if and only if $\phi$ is satisfiable.

\emph{PSPACE-hardness, general case.} Reduce from NFA non-universality, which is
PSPACE-complete \cite{meyer-stockmeyer1972}: given an NFA
$M=(Q,\Sigma,\mu,q_0,\mathrm{Acc})$, decide whether $\mathcal{L}(M)\neq\Sigma^*$. The structure $H$
has a state $z$ with $C(z)=Q$ and $\gamma_0=q_0$, a loop $\ell_a:z\to z$ for each
$a\in\Sigma$ with $R_{\ell_a}(q)=\mu(q,a)$, and a terminal modification
$\#:z\to z_{\mathrm{term}}$ with $C(z_{\mathrm{term}})=\{\ast\}$, $R_\#(q)=\{\ast\}$
if $q\in \mathrm{Acc}$ and $R_\#(q)=\varnothing$ otherwise. A path from $z$ is a word
$w=a_1\cdots a_n$ read by loops, optionally followed by $\#$. By
Proposition~\ref{prop:chains}, the lift after reading $w$ is the set
$\mu^*(q_0,w)$ of states reachable on $w$, and the lift after $w\#$ is
$\{\ast\}$ if $\mu^*(q_0,w)\cap \mathrm{Acc}\neq\varnothing$ and $\varnothing$ otherwise.
The composite relation of a word is the relational composition of its letters'
relations, as functoriality requires.

If $\mathcal{L}(M)\neq\Sigma^*$, take $w\notin \mathcal{L}(M)$; the path $w\#$ gives an empty lift, so
$\gamma_0$ dies. Conversely, suppose $\gamma_0$ dies along some path. The initial
lift $\{q_0\}$ is nonempty. If the lift is empty after reading a prefix $w$ of the
word, then $\mu^*(q_0,w)=\varnothing$, so $w\notin \mathcal{L}(M)$. If it is empty after
$w\#$, then $\mu^*(q_0,w)\cap \mathrm{Acc}=\varnothing$, so again $w\notin \mathcal{L}(M)$. In either
case $\mathcal{L}(M)\neq\Sigma^*$. The construction has size $O(|Q|^2|\Sigma|)$ and is
computable in logarithmic space.

\emph{Complements.} Death-freeness is the complement of death-existence, so it is
coNP-complete for acyclic $H$. For general $H$ it is PSPACE-complete, because
PSPACE is closed under complement.
\end{proof}

\section{Verifications for the Recover\_avail Classification and Related Results}
\label{app:recover-avail-verification}

The acyclic reduction below is for the recoverability problem, not the
death-existence problem of Appendix~\ref{app:complexity-proofs}; the two share the initial fan-out and
variable-layer structure but differ in their fibers and terminal encodings.

This appendix provides the full technical derivations referred to from
Section~\ref{sec:computational-limits} and from Proposition~\ref{prop:avail-not-trace} and Remark~\ref{rem:finite-observation} in Section~\ref{sec:recoverability}.
The instances below use the full-realization baseline of Section~\ref{sec:recoverability}, so realized
and stipulated lifts coincide and lineage death is read off the stipulated lifts
$L_i(\gamma_0)$.

\paragraph{Overview.}
The two reductions are structurally similar to the death-existence
reductions, with the key change that each clause (respectively, each NFA) is
tracked as a \emph{pair} of states: unsatisfied versus satisfied (respectively,
not-yet-accepting versus accepting). The event set $U(\rho)$ consists of the
exclusion events generated when the recognition history excludes the initial
tracked coordinates. \emph{Recover\_avail} then asks whether there exists an
extension along which each excluded coordinate has an eligible descendant that
is realized, undecided, and admissible for inscription. In the acyclic case,
this reduces to CNF satisfiability: the pair-tracking encodes each clause's
satisfaction as a separate terminal reachability condition, and all conditions
must be met along the same assignment-path. In the general case, this reduces
to NFA-intersection-nonemptiness: the machines are simulated in parallel on a
subset of their states over the same input word, and the terminal modification
checks acceptance simultaneously for all machines. The full reductions, with
the layer-by-layer indexing for the acyclic case and the parallel subset
construction for the general case, are given in the subsections below.

\subsection{The acyclic Recover\_avail reduction in detail}

We construct the structure $H$ of Proposition~\ref{prop:recover-avail-classification}
for the acyclic case.

\paragraph{States and modifications.} Let
$\phi = C_1 \wedge \cdots \wedge C_m$ be a CNF formula with $m \geq 1$
clauses over variables $\nu_1, \ldots, \nu_k$, $k\ge1$. The structure $H$ has states
$s_0, s_1, \ldots, s_{k+2}$ and no others, with modifications
$\mathrm{fan}: s_0 \to s_1$ (fan-out),
$\mathrm{true}_i, \mathrm{false}_i: s_i \to s_{i+1}$ for
$i \in \{1, \ldots, k\}$ (variable choices),
and $t: s_{k+1} \to s_{k+2}$ (terminal). The composition condition is
satisfied because the graph is acyclic and composite modifications are
induced by relational composition of the one-step relations.

\paragraph{Fibers and relations.} Let
$C(s_0) = \{\gamma_0\}$.
Let $C(s_1) = \{\xi_1^{(1)}, \ldots, \xi_m^{(1)}\}$ (one clause
continuation per clause, tagged by layer $1$). For $i \in \{2, \ldots, k+1\}$,
let $C(s_i) = \{\mathrm{uns}_j^{(i)}, \mathrm{sat}_j^{(i)} : j = 1, \ldots, m\}$,
where $\mathrm{uns}_j^{(i)}$ denotes ``clause $C_j$ not yet satisfied at layer
$i$'' and $\mathrm{sat}_j^{(i)}$ denotes ``clause $C_j$ satisfied at layer
$i$''. Let $C(s_{k+2}) = \{\delta_1, \ldots, \delta_m\}$.

The fan-out relation is $R_{\mathrm{fan}}(\gamma_0) = \{\xi_1^{(1)}, \ldots, \xi_m^{(1)}\}$.

For each $i \in \{1, \ldots, k\}$, the relations at $\mathrm{true}_i$ and
$\mathrm{false}_i$ act as follows. For each $j \in \{1, \ldots, m\}$:

\begin{itemize}
\item For $i=1$: $R_{\mathrm{true}_1}(\xi_j^{(1)}) = \{\mathrm{sat}_j^{(2)}\}$
  if $\nu_1 = \mathrm{true}$ satisfies clause $C_j$, and
  $R_{\mathrm{true}_1}(\xi_j^{(1)}) = \{\mathrm{uns}_j^{(2)}\}$ otherwise.
  Symmetrically for $\mathrm{false}_1$.
\item For $i \geq 2$: $R_{\mathrm{true}_i}(\mathrm{uns}_j^{(i)}) =
  \{\mathrm{sat}_j^{(i+1)}\}$ if $\nu_i = \mathrm{true}$ satisfies $C_j$, and
  $\{\mathrm{uns}_j^{(i+1)}\}$ otherwise;
  $R_{\mathrm{true}_i}(\mathrm{sat}_j^{(i)}) = \{\mathrm{sat}_j^{(i+1)}\}$
  (once satisfied, stays satisfied). Symmetrically for $\mathrm{false}_i$.
\end{itemize}

In this construction a clause continuation survives when the clause is
satisfied and is carried to $\varnothing$ when it is not, so the surviving
terminal continuations $\delta_j$ correspond to the satisfied clauses. This is
the encoding for the \emph{recoverability} problem, where a path supplies
witnesses for exactly the clauses satisfied along it; the death-existence
reduction of Appendix~\ref{app:complexity-proofs} uses the opposite polarity, carrying a continuation
to $\varnothing$ when the clause is satisfied.

The terminal modification is $R_t(\mathrm{sat}_j^{(k+1)}) = \{\delta_j\}$ and
$R_t(\mathrm{uns}_j^{(k+1)}) = \varnothing$.

\paragraph{Decision domain and criterion.} Let $I_0 = \{\gamma_0\}$ and
$D(s_0) = \{\gamma_0\}$. Let $D(s_1) = C(s_1)$, and let
$S(s_1, \xi_j^{(1)}) = \mathrm{excluded}$ for each $j$. Let
$D(s_{k+2}) = \{\delta_1, \ldots, \delta_m\}$ and
$S(s_{k+2}, \delta_j) = \mathrm{inscribed}$ for each $j$. Let $D$ vanish
at other states.

\paragraph{The recognition history.} Fix the recognition history to exclude
$\xi_j^{(1)}$ at $s_1$ for every $j$ (admissible because $S$ so classifies
them). This generates $m$ unresolved exclusion events
$e_1, \ldots, e_m$ at the prefix $\rho_1$ ending at $s_1$. So
$U(\rho_1) = \{e_1, \ldots, e_m\}$.

\paragraph{The correspondence, forward direction.} A path from $s_0$ to
$s_{k+2}$ is $\mathrm{fan}$, then one choice per variable, then $t$. It encodes a total
assignment $\theta$ of $\nu_1, \ldots, \nu_k$. Suppose $\theta$ satisfies every
clause. Let $p$ be the path that makes the choice matching $\theta$ at each
variable layer. Along $p$, the descendant of $\xi_j^{(1)}$ at $s_{k+2}$ is
$\delta_j$: the clause continuation is carried through the layers
$\mathrm{sat}_j^{(2)}, \ldots, \mathrm{sat}_j^{(k+1)}$ (because $\theta$
satisfies $C_j$ at the layer where the satisfying literal occurs, and once in
$\mathrm{sat}$ it stays there), and the terminal relation sends
$\mathrm{sat}_j^{(k+1)}$ to $\delta_j$. At the prefix $\eta = p|_{s_{k+2}}$,
the state is $s_{k+2}$, the coordinate $\delta_j$ has just entered the decision
domain and is undecided, and $\widehat S(s_{k+2}, \delta_j) = \mathsf I$. So
$(\eta, \delta_j)$ is a corrective-route witness for $e_j$. Since this holds
for every $j$, all $m$ events have witnesses on the same extension, and
$\operatorname{Recover}_{\mathrm{avail}}(\rho_1)$ holds.

\paragraph{The correspondence, reverse direction.} Suppose
$\operatorname{Recover}_{\mathrm{avail}}(\rho_1)$ holds. Then there is an
extension $\rho' \succeq \rho_1$ and, for each $j$, a prefix $\eta_j$ with
$\rho_1\preceq\eta_j\preceq\rho'$ and a witness $(\eta_j, \delta^{(j)})$ for
$e_j$; the definition gives each event its own prefix, and we show that they
coincide. The witness requires
$\delta^{(j)} \in R_{p_j}(\xi_j^{(1)}) \cap D(y_j)$, where $p_j$ is the segment
of $\rho'$ from $s_1$ to the state $y_j$ at which $\eta_j$ ends. Since $D$
vanishes at every state except $s_0, s_1, s_{k+2}$, the candidates are
$y_j\in\{s_0,s_1,s_{k+2}\}$. No modification enters $s_0$, so $y_j\neq s_0$. The
state $s_1$ is reached only at $\rho_1$, the end of the given history, so the only
candidate there is $\eta_j=\rho_1$ with the zero-length segment, for which
$R_{p_j}(\xi_j^{(1)})\cap D(s_1)=\{\xi_j^{(1)}\}$. That coordinate was decided
$\mathsf E$ in $\rho_1$, so $B_{\eta_j}(s_1)(\xi_j^{(1)})\neq\mathsf U$ and
$\widehat S=\mathsf E\neq\mathsf I$: it is not a witness. The graph is acyclic,
so $s_{k+2}$ is reached at most once, at the end of any extension. Hence
$y_j = s_{k+2}$ and $\eta_j$ is the end of $\rho'$ for every $j$: the prefixes
coincide, and every witness uses the same segment $p$ of $\rho'$ from $s_1$ to
$s_{k+2}$. Moreover $\delta^{(j)} \in
C(s_{k+2}) = \{\delta_1, \ldots, \delta_m\}$. Trace the relations: the only way
for a clause continuation to reach any $\delta_\ell$ is via the terminal
relation $t$, which sends $\mathrm{sat}_\ell^{(k+1)}$ to $\delta_\ell$; the
only way to reach $\mathrm{sat}_\ell^{(k+1)}$ is by staying in the
$\mathrm{sat}$ chain from the layer at which the clause first becomes
satisfied. Hence $\delta^{(j)} \in R_p(\xi_j^{(1)})$ implies
$\delta^{(j)} = \delta_j$ and the path $p$ satisfies $C_j$. This holds for
every $j$, so the assignment $\theta$ encoded by the variable choices along
$p$ satisfies every clause. Hence $\phi$ is satisfiable. No alternative route
can supply a witness for $e_j$: the only state where $D$ is nonempty and a
$\delta$-type coordinate is reachable is $s_{k+2}$, and the only relation
producing $\delta_j$ is $t$ applied to $\mathrm{sat}_j^{(k+1)}$.

The only transition from the $\mathrm{uns}$ chain to the $\mathrm{sat}$ chain
is at the layer where a selected literal satisfies that clause; no relation at
any layer maps $\mathrm{uns}_j^{(i)}$ directly to a $\mathrm{sat}$ coordinate
of another clause, or to $\mathrm{sat}_j^{(i')}$ for $i' > i+1$. So a clause
that has not been satisfied at any layer remains in the $\mathrm{uns}$ chain
through every subsequent layer, and its terminal image under $t$ is empty. A
satisfied clause reaches exactly $\delta_j$.

Hence $\operatorname{Recover}_{\mathrm{avail}}(\rho_1)$ holds iff there is an
assignment $\theta$ satisfying every clause, that is, iff $\phi$ is
satisfiable.

\paragraph{NP membership.} A satisfying assignment is a certificate of
polynomial size. Under the polynomial-time evaluation restriction stated in
Section~\ref{sec:computational-limits}, verification takes polynomial time:
propagate the clause continuations through the chosen assignment and check that
every $\delta_j$ is in the terminal lift, in $D(s_{k+2})$, and satisfies
$\widehat S(s_{k+2}, \delta_j) = \mathsf I$.

This completes the acyclic reduction.

\subsection{The general Recover\_avail reduction in detail}

We construct the structure $H$ of Proposition~\ref{prop:recover-avail-classification}
for the general case, reducing from NFA-intersection-nonemptiness.

\paragraph{Input.} Let $A_1, \ldots, A_k$ be NFAs over a common alphabet
$\Sigma$, with state sets $Q_1, \ldots, Q_k$, transition relations
$\mu_i: Q_i \times \Sigma \to 2^{Q_i}$, initial states $q_{0,i}$, and
accepting sets $\mathrm{Acc}_i$. The question is whether
$\bigcap_{i=1}^k \mathcal{L}(A_i) \neq \varnothing$. This problem is
PSPACE-complete \cite{kozen1977}.

\paragraph{States and fibers.} The structure has states
$z_0, z_1, z_2$. Let $C(z_0) = \{\gamma_0\}$. Let
$C(z_1) = Q_1 \sqcup \cdots \sqcup Q_k$ be the disjoint union of the state
sets, with elements tagged by machine index so that blocks are disjoint. Let
$C(z_2) = \{\mathrm{ACCEPT}_1, \ldots, \mathrm{ACCEPT}_k\}$.

\paragraph{Modifications.} The fan-out modification $\mathrm{fan}: z_0 \to z_1$ has
$R_{\mathrm{fan}}(\gamma_0) = \{q_{0,1}, \ldots, q_{0,k}\}$. For each $a \in \Sigma$,
the loop modification $\ell_a: z_1 \to z_1$ has
$R_{\ell_a}(q) = \mu_i(q, a)$ for $q \in Q_i$, block-wise. The terminal
modification $t: z_1 \to z_2$ has $R_t(q) = \{\mathrm{ACCEPT}_i\}$ if
$q \in \mathrm{Acc}_i$ (with $q \in Q_i$), and $R_t(q) = \varnothing$ otherwise, again
block-wise.

\paragraph{Composition condition.} The composition of loop modifications
$\ell_a$ and $\ell_b$ acts on $z_1$; $R_{\ell_a \circ \ell_b}(q)$ for
$q \in Q_i$ equals $R_{\ell_a}(R_{\ell_b}(q)) = \mu_i(\mu_i(q, b), a)$,
which is the NFA transition for reading the two-letter word $ba$. Blocks
remain disjoint, so composition holds block-wise. The composite
$t \circ \ell_a$ acts by $R_t(R_{\ell_a}(q))$, matching the NFA transition
followed by the terminal check. The empty set is absorbing under relational
composition.

\paragraph{Decision domain and criterion.} Let $I_0 = \{\gamma_0\}$ and
$D(z_0) = \{\gamma_0\}$. Let
$D(z_1) = \{q_{0,1}, \ldots, q_{0,k}\} \subsetneq C(z_1)$, and let
$S(z_1, q_{0,i}) = \mathrm{excluded}$ for each $i$. Let
$D(z_2) = \{\mathrm{ACCEPT}_1, \ldots, \mathrm{ACCEPT}_k\}$ and
$S(z_2, \mathrm{ACCEPT}_i) = \mathrm{inscribed}$ for each $i$. Let $D$
vanish at other states.

\paragraph{The recognition history.} Fix the recognition history to exclude
$q_{0,i}$ at $z_1$ for every $i$ (admissible because $S$ so classifies
them). This generates $k$ unresolved exclusion events
$e_1, \ldots, e_k$ at the prefix $\rho_1$ ending at $z_1$. So
$U(\rho_1) = \{e_1, \ldots, e_k\}$.

\paragraph{The correspondence, forward direction.} A path from $z_0$ to $z_2$
is $\mathrm{fan}$, then a finite sequence of loop modifications $\ell_{a_1}, \ldots,
\ell_{a_n}$ (possibly empty, corresponding to the empty word), then $t$. It
encodes a word $w = a_1 \cdots a_n \in \Sigma^*$.

Suppose $w \in \bigcap_i \mathcal{L}(A_i)$. For each $i$, the lift of $q_{0,i}$ after
reading $w$ is $\mu_i^*(q_{0,i}, w)$, which contains some state in $\mathrm{Acc}_i$
(since $w \in \mathcal{L}(A_i)$). The terminal relation sends that state to
$\mathrm{ACCEPT}_i$, so $\mathrm{ACCEPT}_i \in R_t(\mu_i^*(q_{0,i}, w))$.
At the prefix $\eta = p|_{z_2}$, the state is $z_2$, the coordinate
$\mathrm{ACCEPT}_i$ has just entered the decision domain and is undecided, and
$\widehat S(z_2, \mathrm{ACCEPT}_i) = \mathsf I$. So $(\eta, \mathrm{ACCEPT}_i)$
is a corrective-route witness for $e_i$. Since this holds for every $i$, all
$k$ events have witnesses on the same extension, and
$\operatorname{Recover}_{\mathrm{avail}}(\rho_1)$ holds.

The construction is a parallel subset simulation of the $k$ machines on the
same word, not a product of their individual states. The fiber
$C(z_1) = Q_1 \sqcup \cdots \sqcup Q_k$ is the disjoint union of the state
sets, and the fan-out $R_{\mathrm{fan}}(\gamma_0) = \{q_{0,1}, \ldots, q_{0,k}\}$ places
the initial state of every machine in the lift simultaneously. The loop
modifications act block-wise, and the transition relation $\mu_i$ is
set-valued, so at each step the lift is a subset of the disjoint union
containing a subset of each block. For the path $p$ that encodes $w$,
\[
R_p(\{q_{0,i}\}) = \mu_i^*(q_{0,i}, w),
\]
the full set of states reachable by $A_i$ on $w$; the machine may be in
several states at once, or in none, and the lift tracks all of them
simultaneously.

The terminal modification $t: z_1 \to z_2$ maps every accepting state in
block $i$ to $\mathrm{ACCEPT}_i$ and every non-accepting state to
$\varnothing$. Hence
\[
\mathrm{ACCEPT}_i \in R_t(R_p(\{q_{0,i}\})) \iff
\mu_i^*(q_{0,i}, w) \cap \mathrm{Acc}_i \neq \varnothing \iff w \in \mathcal{L}(A_i).
\]
Therefore $e_i$ has a witness on this path iff $w \in \mathcal{L}(A_i)$, and all
exclusion events have witnesses on the same path exactly when some word is
accepted by every $A_i$.

\paragraph{The correspondence, reverse direction.} Conversely, suppose
$\operatorname{Recover}_{\mathrm{avail}}(\rho_1)$ holds. Then there is an
extension $\rho' \succeq \rho_1$ and, for each $i$, a prefix $\eta_i$ with
$\rho_1\preceq\eta_i\preceq\rho'$ and a witness $(\eta_i, \delta^{(i)})$ for
$e_i$; the definition gives each event its own prefix, and we show below that
they coincide. The witness requires
$\delta^{(i)} \in R_{p_i}(\{q_{0,i}\}) \cap D(y_i)$ where $p_i$ is the segment of
$\rho'$ from $z_1$ to the state $y_i$ at which $\eta_i$ ends.

Since $D$ vanishes at every state except $z_0, z_1, z_2$, the candidates are
$y_i\in\{z_0,z_1,z_2\}$. No modification enters $z_0$, so $y_i\neq z_0$. The state
$z_1$ is revisited by the loops $\ell_a$, so it is not excluded by position; it
is excluded by the data. The loops and $R_{\mathrm{fan}}$ act block-wise, so
$R_{p_i}(\{q_{0,i}\})\subseteq Q_i$, and $D(z_1)\cap Q_i=\{q_{0,i}\}$. The only
possible witness at $z_1$ is therefore the coordinate $q_{0,i}$ itself, whether
at $\eta_i=\rho_1$ (zero-length segment) or after some loops, which
has $B=\mathsf E$ from $\rho_1$ (no transition alters a completed coordinate)
and $\widehat S=\mathsf E\neq\mathsf I$; it is not undecided and not
admissible for inscription. Hence $y_i = z_2$ and
$\delta^{(i)} \in C(z_2) = \{\mathrm{ACCEPT}_1, \ldots, \mathrm{ACCEPT}_k\}$. Since
$\rho'$ reaches $z_2$ at most once and no modification leaves $z_2$, the state
$z_2$ is reached only at the end of $\rho'$, so all $\eta_i$ coincide with it and
every witness uses the same segment $p$ of $\rho'$ from $z_1$ to $z_2$.

From $z_1$, the only modifications available are the loops $\ell_a$ and the
terminal $t: z_1 \to z_2$. A path from $z_1$ to $z_2$ is therefore a finite
sequence of loop modifications followed by $t$, i.e., $p = t \circ \ell_{a_n} \circ \cdots \circ \ell_{a_1}$ for some word $w = a_1 \cdots a_n$.

Then $R_p(\{q_{0,i}\}) = R_t(R_{\ell_{a_n} \circ \cdots \circ \ell_{a_1}}
(\{q_{0,i}\})) = R_t(\mu_i^*(q_{0,i}, w))$, which is
$\{\mathrm{ACCEPT}_i\}$ if $\mu_i^*(q_{0,i}, w) \cap \mathrm{Acc}_i \neq \varnothing$
and $\varnothing$ otherwise. So a witness for $e_i$ exists iff
$w \in \mathcal{L}(A_i)$. The segment $p$ is common to all events, as shown
above, and encodes a single word $w$. Hence a single word
$w$ is accepted by every $A_i$.

No other terminal coordinate can witness event $e_i$: the terminal relation is
block-wise, so for $q \in Q_i$ the image $R_t(q)$ is either
$\{\mathrm{ACCEPT}_i\}$ or $\varnothing$, never a terminal coordinate of
another block. The witness must use the same extension and hence the same word
for every event.

Hence $\operatorname{Recover}_{\mathrm{avail}}(\rho_1)$ holds iff
$\bigcap_i \mathcal{L}(A_i) \neq \varnothing$.

\paragraph{PSPACE membership.} The argument is given for the restricted,
state-determined model of Section~\ref{sec:computational-limits}: the
determination rule $\Delta$ is a function of the state reached alone, and is
evaluable in space polynomial in the extensional input.
Let $\rho$ be the given history. The search is written directly from
Eq.~\eqref{eq:recover-exists}, which asks for one extension $\rho'$ that is
death-free and gives every event of $U(\rho)$ a witness.

\emph{Fixed data.} The recognition state $B_\rho$, the set $U(\rho)$ with the
excluded coordinate $\gamma_e$ and state $x_e$ of each event, and $I_0$ are
computed from $\rho$ and do not change during the search. If $U(\rho)$ contains
a lineage-death event, the instance is rejected, since such an event has no
witness.

\emph{Configuration.} A configuration is
$c=\bigl(y,\,(L_e)_{e\in U(\rho)},\,(M_{\gamma_0})_{\gamma_0\in I_0},\,(b_e)_{e\in U(\rho)}\bigr)$:
the current state $y$ of the walk; for each unresolved event $e$, the lift
$L_e=R_p(\gamma_e)\subseteq C(y)$ along the segment $p:x_e\leadsto y$ of the
walk so far; for each $\gamma_0\in I_0$, its current lift
$M_{\gamma_0}\subseteq C(y)$; and one bit $b_e$ per event, indexed by the event,
recording whether a witness for $e$ has been found. The initial configuration
is computed from $\rho$. A configuration has size
$O\bigl((|U(\rho)|+|I_0|)\cdot m+|U(\rho)|+\log|X|\bigr)$ bits, where $m$ is
the largest fiber, which is polynomial in the input.

\emph{Transition.} A step chooses a modification $f:y\to y'$ and replaces every
$L_e$ and $M_{\gamma_0}$ by its image under $R_f$, computed with $\Delta$; all
events thus advance along the same walk. It then sets $b_e:=1$ if $b_e=1$
already, or if some $\delta\in L_e\cap D(y')$ has $\widehat S(y',\delta)=\mathsf I$
and is undecided at the current prefix. The same test is applied once to the
initial configuration, for $\eta=\rho$.

\emph{Acceptance.} Accept at the first configuration in which every $b_e=1$,
provided no $M_{\gamma_0}$ has been empty at any configuration so far.

Each clause of Eq.~\eqref{eq:recover-exists} is carried by one feature of this
search. \emph{Common extension:} there is one walk, and every $L_e$ is updated
by every step; the bits are indexed by event, so each event may be witnessed at
its own prefix of that walk, as the definition allows. \emph{Death-freeness:}
the definition asks for no death event in $F(\rho')$, which also excludes a
witness that is followed by a later death. If $\rho'$ is a witnessing extension,
its truncation at the last witnessing prefix is also one: each witness depends
only on the history up to its prefix $\eta$, and $F$ is monotone, so truncation
cannot introduce a death event. The search stops at the first configuration with
every bit set, which is that truncation, so checking for death up to that point
suffices. \emph{Undecidedness at the witness prefix:} the condition
$B_\eta(y)(\delta)=\mathsf U$ is evaluated at the prefix $\eta$ at which the
witness is found. A recognition event beyond $\rho$ replaces $\mathsf U$ by a
value and so cannot create a witness; it is enough to search extensions with no
such events. For a coordinate at a state reached by $\rho$, the value at $\eta$
is then its value in $B_\rho$, and a coordinate first reached beyond $\rho$ is
initialized to $\mathsf U$. The test above therefore reads only the fixed table
$B_\rho$.

The number of configurations is at most
$|X|\cdot 2^{(|U(\rho)|+|I_0|)m}\cdot 2^{|U(\rho)|}$, exponential in the input.
A walk that revisits a configuration can be shortened by deleting the loop,
which loses no witness bit, since equal configurations carry equal bits. A
nondeterministic machine can therefore guess the walk one modification at a
time, storing only the current configuration and a step counter bounded by the
number of configurations, which takes polynomial space. Evaluating $\Delta$ at
each step uses polynomial space, so the weaker polynomial-space assumption
suffices for membership; the polynomial-time assumption is needed only for the
acyclic NP upper bound. The problem is in $\mathrm{NPSPACE}$, and by Savitch's
theorem~\cite{savitch1970} $\mathrm{NPSPACE}=\mathrm{PSPACE}$, so
$\operatorname{Recover}_{\mathrm{avail}}$ is in PSPACE under the restricted
model.

Death-freeness is part of the configuration above, through the lifts
$M_{\gamma_0}$. The reductions are unaffected; see the death-freeness
observation at the end of this subsection.

\paragraph{On the two evaluability assumptions.}
The acyclic reduction uses the polynomial-time evaluability of $\Delta$ to
verify a path certificate: the certificate is the path, and the verifier
computes the lift step by step, applying $\Delta$ once per step. The general
reduction uses only polynomial-space evaluability, because the certificate is
a walk with a counter and the verifier simulates the walk rather than
outputting it. The two assumptions are not different in kind; they are the
same requirement at different strengths, and the choice is dictated by the
certificate type. A path certificate could also be verified by simulation,
which would weaken the acyclic bound to PSPACE without changing the result,
since NP is contained in PSPACE. The stronger assumption is stated for the
acyclic case because the NP bound is the sharp one for that case, and the
sharp bound requires the certificate to be checkable in polynomial time.

\paragraph{Death-freeness of the witnessing extensions.}
The death-freeness conjunct of Eq.~\eqref{eq:recover-exists} is satisfied
automatically along the witnessing extension in both reductions. In each, every
excluded coordinate is a direct image of $\gamma_0$ under the fan-out
($\xi_j^{(1)}$, respectively $q_{0,i}$); a witness for it lies in $R_p$ of that
coordinate and so in the lift of $\gamma_0$ at the terminal state ($s_{k+2}$,
respectively $z_2$), where $\rho'$ ends; and a lift that is empty at some stage
is empty at every later stage, so $L_i(\gamma_0)\neq\varnothing$ at every stage
of $\rho'$. The observation is used in the forward directions and in the
membership arguments, where the configuration tracks the lift of $\gamma_0$; the
reverse directions use only the witness. It is specific to these constructions:
in a general instance an excluded coordinate need not descend from $\gamma_0$,
and a witnessing extension need not be death-free.

\subsection{Verification of the availability versus trace proposition}
\label{app:avail-vs-trace}

\paragraph{Positive direction (strong fairness).} Suppose the structure has
finitely many coordinates, the recognition history is strongly fair, $\tau$
contains no lineage-death event, and for each exclusion event $e\in F(\tau)$ a
witness for $e$ keeps occurring along $\tau$ for as long as $e$ is unresolved.
Fix $e$ and suppose, for contradiction, that $e$ is never resolved:
$e\in U(\tau|i)$ for all $i\ge i_0$. By hypothesis, for every $i\ge i_0$ there
are $j\ge i$ and a witness $(\tau|j,\delta_j)$ for $e$, ending at a state $y_j$.
Since there are finitely many coordinates, some coordinate $(y,\delta)$ is the
witness coordinate at infinitely many $j$. At each such $\tau|j$ the coordinate
is undecided, and a completed coordinate is never altered, so it is undecided
at every prefix of $\tau$. At each such $j$ the transition
$\mathsf U\to\mathsf I$ at $(y,\delta)$ is admissible: the witness gives
$\delta\in R_p(\gamma)\cap D(y)$, $B=\mathsf U$ and
$\widehat S(y,\delta)=\mathsf I$. The transition is thus admissible infinitely
often, so by strong fairness it is eventually taken, and the coordinate is then
no longer undecided, contrary to the previous sentence. Hence $e$ is resolved.
Since $e$ was arbitrary and $\tau$ contains no death event,
$\operatorname{Recover}_{\mathrm{trace}}(\tau)$ holds. The hypothesis requires a
cycle through the opportunity state, so this direction never applies to acyclic
structures, where every history is finite. The hypothesis is stated for as long
as $e$ is unresolved, and not as infinite recurrence outright, because a witness
requires the coordinate to be undecided: on a strongly fair history a witness
cannot recur infinitely often, since fairness forces the transition and ends
the opportunity.

\paragraph{Negative direction.} We construct the infinite history $\tau$ of
Proposition~\ref{prop:avail-not-trace}.

\paragraph{Setup.} Fix a finite structure $H$ containing a state $y$
with the following property: at $y$ there is a cycle
$\ell: y \to y$ with $R_\ell(\delta) = \{\delta\}$ for some
$\delta \in C(y)$, and $\delta$ is a structurally eligible descendant of a
coordinate $\gamma$ that has been excluded at an earlier prefix. Suppose
$\delta$ also satisfies $\delta \in D(y)$ and
$\widehat S(y, \delta) = \mathsf I$. Write $e=(\rho_e,\gamma)$ for the exclusion event of $\gamma$ at a state $x$
that $\tau$ visits before its first arrival at $y$, and suppose $e$ is the only
failure event generated along $\tau$ (in particular, no lineage-death event
occurs).

\paragraph{The infinite history.} Let $\tau$ be the infinite history that,
from some point on, revisits $y$ by applying $\ell$ at each step. Let the
recognition history never perform the transition $\mathsf U \to \mathsf I$ at
$(y, \delta)$, though the transition is admissible at every visit.

\paragraph{Availability at every prefix.} Fix any finite prefix
$\tau|i$ of $\tau$. Since $\tau$ revisits $y$ infinitely often, there is $j\ge i$ such that $\tau|j$ ends at $y$. Take $\rho'=\eta=\tau|j$. The
segment $p$ from the state $x$ of the exclusion to $y$ along $\tau|j$ carries
$\delta$ in $R_p(\gamma)$: $\delta$ is a structurally eligible descendant at
the first arrival at $y$, and each further application of $\ell$ preserves it
because $R_\ell(\delta)=\{\delta\}$. No modification out of $y$ is needed,
since the witness sits at $y$ itself. Because the recognition history never
performs the transition at $(y,\delta)$, the pair $(\eta,\delta)$ satisfies
\[
\delta \in R_p(\gamma) \cap D(y),\qquad
B_{\eta}(y)(\delta) = \mathsf U,\qquad
\widehat S(y, \delta) = \mathsf I,
\]
so it is a corrective-route witness for $e$. Hence
$W_e(\rho') \neq \varnothing$, and $\operatorname{Recover}_{\mathrm{avail}}(\tau|i)$
holds.

\paragraph{Trace failure.} Under the stipulated recognition history, the
transition $\mathsf U \to \mathsf I$ at $(y, \delta)$ is never performed. So
for every prefix $\tau|i$ with $e\in F(\tau|i)$, also $e \in U(\tau|i)$. Hence
$\operatorname{Recover}_{\mathrm{trace}}(\tau)$ fails, because $e$ never
leaves $U$.

\paragraph{Consequence.} The predicate
$\operatorname{Recover}_{\mathrm{avail}}$ can hold at every finite prefix of
an infinite history while $\operatorname{Recover}_{\mathrm{trace}}$ fails on
that same history. The failure is not epistemic --- it is a consequence of the
framework's separation of admissibility from occurrence, and with a strongly
fair recognition history it does not arise.

\subsection{Verification of the constructed asymmetry in Remark~\ref{rem:finite-observation}}

Assume the structure contains neutral modifications that can be repeated indefinitely. We construct the two infinite extensions used in the remark.

Fix a finite prefix $\rho$ ending with an unresolved exclusion event
$e = (\rho_e, \gamma)$. Suppose $\gamma$ has a structurally eligible
descendant $\delta$ at a state $y$ reachable from the end of $\rho$ by a
finite path $p$, with $\delta \in D(y)$ and
$\widehat S(y, \delta) = \mathsf I$.

\paragraph{Neutral extensions.} Let $n \geq 1$. From $\rho$, extend by $n$
neutral modifications that do not affect the structural image of $\gamma$
or the structure of $p$; that is, they are modifications on states disjoint
from $y$ and from the path $p$, or self-loops on states whose fibers do not
carry $\gamma$-descendants. Let $\rho^{+n}$ denote the resulting prefix of
length $n$ beyond $\rho$.

\paragraph{Two extensions.} From $\rho^{+n}$, define:

\begin{itemize}
\item $\tau_1$: extend by the path $p$ to $y$, and at $y$ realize and inscribe
  $\delta$. Then $e$ is resolved on $\tau_1$.
\item $\tau_2$: extend by a modification whose relation carries the image of
  $\gamma$ to $\varnothing$ on every branch from the current state, and then
  by any further modifications. Then $e$ is never resolved on $\tau_2$.
\end{itemize}

\paragraph{Prefix-indistinguishability.} Since both $\tau_1$ and $\tau_2$
agree with $\rho^{+n}$ on the first $n$ steps beyond $\rho$, and $n$ is
arbitrary, no finite prefix of $\rho$ determines whether $e$ is eventually
resolved. This establishes the asymmetry claimed in the remark.

\section{Realization}
\label{app:realization}

The abstract model of Section~\ref{sec:model} gives a stipulated structure. This appendix adds
conformance determination: which of its continuations are found to be realized,
distinguishing non-realization from a realized continuation that is set aside.

\paragraph{Realization is history-indexed.}
The stipulated continuation structure is fixed: $C(x)$ and $R_f$ depend on the
state and the modification but not on the history by which the state was
reached. Realization is not: the realized subset $C^{\mathrm{re}}_\rho(x)$ is
attached to the pair $(x,\rho)$---a state together with the finite history
reaching it---because the conformance evidence available at the end of a
history may depend on the history, not merely on the state. The same structural
state can therefore have different realized subsets on different histories. The
recoverability predicates of Section~\ref{sec:recoverability} respect this indexing by quantifying
over extensions of a given history.
The complexity results of Section~\ref{sec:computational-limits} are stated
for a restricted model in which $\Delta$ is state-determined and effectively
evaluable; the general definition above permits history-indexed realization,
which is not covered by those results.

\subsection{Setting}

For a system without explicit compositional structure, the pair $(C,R)$ is
\emph{stipulated}. Stipulation is a governance act. Three things are separated:
a structural fact is a consequence of the stipulation; a conformance
determination records which stipulated continuations the system realizes; a
recognition record operation inscribes or excludes. The first is formal. The
second and third are accountable acts and are recorded as such.

In this appendix, we use the abbreviation
$C^{\mathrm{re}}_i=C^{\mathrm{re}}_\rho(x_i)$ for the realized subset at stage
$i$ of the realized history $\rho$. A recognition record is \emph{realized} if
it additionally satisfies
\[
I(x)\subseteq D^{\mathrm{re}}_\rho(x)\subseteq C^{\mathrm{re}}_i\subseteq C(x)
\]
at each realized coordinate, with $D^{\mathrm{re}}_\rho(x)=D(x)\cap C^{\mathrm{re}}_i$.
The main recognition record of Section~\ref{sec:decision} is defined only by
$I(x)\subseteq D(x)$. The realized condition applies when recognition is
restricted to realized continuations, which is the case of interest in this
section. If every continuation is realized, the chain reduces to
$I(x)\subseteq D(x)\subseteq C(x)$.

\paragraph{Non-realization as the encoding of feature loss.}
A failed transformation and a non-realized continuation are different events.
A failed transformation may prevent the successor state from being reached at
all, whereas a non-realized continuation presupposes that a state and its
continuation space are available for determination. The realization-cut
analysis assumes that the relevant history and its state layers have been
specified, and represents loss of a continuation by non-realization of the
corresponding vertex in the path graph. Failures that prevent the history
itself from proceeding require a different encoding and lie outside the graph
model developed in this appendix. The realization-cut number introduced below
measures robustness against non-realization within a stipulated path graph.

\paragraph{Determinations can be revised.}
The determination rule $\Delta$ is fixed within a single analysis. A later
determination may revise an earlier one, for example when new conformance
evidence becomes available. The framework treats determinations as fixed for
the duration of an analysis; a revision would be modeled by re-running the
analysis with a different $\Delta$ and recording the change as a new
determination. The framework records determinations and their provenance; it
does not adjudicate between competing determinations.

\subsection{Realized subsets and realized lifts}

Fix a determination rule $\Delta$ whose input is a finite generating-history
prefix $\pi$ ending at state $x$ together with the conformance evidence recorded
for that prefix, and whose output is a subset $\Delta(\pi)\subseteq C(x)$. The
rule is fixed across all extensions. In the general model the rule is an
exogenous input: the framework does not constrain what $\Delta$ may depend on,
and no complexity claim is made for it. The restriction used for the
recoverability classifications --- state-determined and effectively evaluable
--- is stated in Section~\ref{sec:computational-limits} and repeated where it
is invoked.

A finite realized history $\rho$ consists
of a generating-history prefix $\pi=(f_1,\dots,f_n)$, with states
$x_0,\dots,x_n$, together with the realized subsets assigned by $\Delta$ to its
prefixes. We stipulate $\Delta(\pi|0)=C(x_0)$ at the initial prefix. Write
$C^{\mathrm{re}}_i=\Delta(\pi|i)\subseteq C(x_i)$. The stipulation
$C^{\mathrm{re}}_0=C(x_0)$ is an initial-realization convention.

Given $\gamma_0\in C(x_0)$, the \emph{realized lifts} are
\begin{equation}
L^{\mathrm{re}}_0(\gamma_0)=\{\gamma_0\},\qquad
L^{\mathrm{re}}_i(\gamma_0)=R_{f_i}\bigl(L^{\mathrm{re}}_{i-1}(\gamma_0)\bigr)\cap C^{\mathrm{re}}_i .
\label{eq:realized-lift}
\end{equation}
For a path segment $p$ along $\rho$, we write $R^{\mathrm{re}}_p(\gamma)$ for
the realized lift obtained by the same recursion along that segment, including
intersection with the realized subset at each intermediate state. The statuses
$\mathrm{Dead}^{\mathrm{re}}$, $\mathrm{Pres}^{\mathrm{re}}$,
$\mathrm{Dil}^{\mathrm{re}}$ are defined from $L^{\mathrm{re}}_i$ exactly as
the stipulated statuses are defined from $L_i$.

\emph{Realized death} is $L^{\mathrm{re}}_i(\gamma_0)=\varnothing$ for some
stage $i$.

\begin{remark}[Realization cannot create stipulated branches]
\label{rem:realization}
For every $\gamma_0$ and every finite realized history $\rho$,
$L^{\mathrm{re}}_i(\gamma_0)\subseteq L_i(\gamma_0)\cap C^{\mathrm{re}}_i$ for
all $i$, so $|L^{\mathrm{re}}_i(\gamma_0)|\le|L_i(\gamma_0)|$: realization
cannot create a branch absent from the stipulated structure. If $\gamma_0$ is
$\mathrm{Dead}$ along $\rho$ under the stipulated structure then it is
$\mathrm{Dead}^{\mathrm{re}}$, and if it is $\mathrm{Pres}$ then it is
$\mathrm{Pres}^{\mathrm{re}}$ or $\mathrm{Dead}^{\mathrm{re}}$. A diluted
lineage may be realized as diluted, preserved, or dead: if all but one of its
stipulated branches are unrealized, the realized lift has cardinality one at
every stage, so the lineage is $\mathrm{Pres}^{\mathrm{re}}$ even though it is
$\mathrm{Dil}$ in the stipulated structure. The inclusion is by induction:
$L^{\mathrm{re}}_0=L_0$, and if
$L^{\mathrm{re}}_{i-1}\subseteq L_{i-1}$ then monotonicity of $R_{f_i}$ on
subsets gives
$L^{\mathrm{re}}_i=R_{f_i}(L^{\mathrm{re}}_{i-1})\cap C^{\mathrm{re}}_i\subseteq R_{f_i}(L_{i-1})\cap C^{\mathrm{re}}_i=L_i\cap C^{\mathrm{re}}_i$.
\end{remark}

\begin{remark}[No composition law for realized relations]
Composition is a property of the stipulated $R$. The realized lift passes
through the realized subset at each intermediate state, so the realized
relation along a path can be strictly smaller than the restriction of the
composite stipulated relation to realized endpoints. The identification
$L_n=R_{f_n\circ\cdots\circ f_1}$ of Eq.~\eqref{eq:iterated-composite} holds for the
stipulated structure only.
\end{remark}

\subsection{Realization-cut number}

Fix a finite realized history $\rho=(f_1,\dots,f_n)$ with $n\ge1$ and a
continuation $\gamma_0\in C(x_0)$ that is not stipulated-dead along $\rho$.
Assume the continuation fibers along $\rho$ are finite, so that the path graph
defined below is finite. The \emph{path graph} $G_\rho(\gamma_0)$ has as
vertices the pairs $(i,\varepsilon)$ with $\varepsilon\in L_i(\gamma_0)$, and
an edge $(i-1,\varepsilon)\to(i,\varepsilon')$ whenever
$\varepsilon'\in R_{f_i}(\varepsilon)$. Layer $i$ is
$\{i\}\times L_i(\gamma_0)$. Every edge joins consecutive layers, so every path
from $(0,\gamma_0)$ to layer $n$ contains exactly one vertex of each layer. A
vertex $(i,\varepsilon)$ is \emph{realized} if
$\varepsilon\in C^{\mathrm{re}}_i$. By the recursion in
Eq.~\eqref{eq:realized-lift}, an element lies in
$L^{\mathrm{re}}_i(\gamma_0)$ if and only if it is realized and has a
predecessor in $L^{\mathrm{re}}_{i-1}(\gamma_0)$. Hence
$L^{\mathrm{re}}_i(\gamma_0)$ is the set of layer-$i$ vertices reachable from
$(0,\gamma_0)$ by a path all of whose vertices are realized.

Let $\kappa_n(\gamma_0)$ be the least cardinality of a set $K$ of vertices of
$G_\rho(\gamma_0)$, excluding $(0,\gamma_0)$, such that declaring exactly the
vertices in $K$ unrealized disconnects $(0,\gamma_0)$ from layer $n$; we call
it the \emph{realization-cut number} of $\gamma_0$ along $\rho$. The quantity
is defined against a fully realized baseline, and it measures the minimum
number of non-realizations required to empty the realized lift. Equivalently,
it is a vertex-connectivity measure from the initial continuation to the final
layer: larger values mean that more distinct continuation vertices must be
made unrealized to block every stipulated path. It measures worst-case
combinatorial robustness, not the probability of failure or the behavior of a
particular determination process.

The persistence statuses of Section~\ref{sec:persistence} classify a lineage
as preserved, diluted, or dead. That classification is coarse: two diluted
lineages with different branch structures receive the same status. The cut
number refines the classification. Among diluted lineages, it measures how
many independent realization failures the lineage can survive before the
realized lift empties, so it is the robustness counterpart of the branching
classification. A preserved lineage has $\kappa_n=1$, since a unique path to
the terminal layer can be blocked by a single unrealized vertex; a diluted
lineage has $\kappa_n\le\min_i|L_i(\gamma_0)|$, so a singleton layer anywhere
in layers $1$ through $n$ caps the cut number at $1$ however wide the other
layers are. The statuses say whether the lineage branches; the cut number says
how much redundancy the branching provides.

\begin{proposition}[Realization-cut number]
\label{prop:robustness}
Assume the continuation fibers along $\rho$ are finite, so that
$G_\rho(\gamma_0)$ is a finite directed graph. With the notation above:
\begin{enumerate}[label=\textnormal{(\alph*)}]
\item $L^{\mathrm{re}}_n(\gamma_0)=\varnothing$ if and only if the set of
unrealized vertices of $G_\rho(\gamma_0)$ meets every path from
$(0,\gamma_0)$ to layer $n$; equivalently, $\gamma_0$ is
$\mathrm{Dead}^{\mathrm{re}}$ along $\rho$.
\item $\kappa_n(\gamma_0)$ equals the maximum number of paths from
$(0,\gamma_0)$ to layer $n$ that are pairwise vertex-disjoint except at
$(0,\gamma_0)$.
\item $1\le\kappa_n(\gamma_0)\le\min_{1\le i\le n}|L_i(\gamma_0)|$.
\item $\kappa_m(\gamma_0)\ge\kappa_n(\gamma_0)$ for $1\le m\le n$: the cut
number does not increase as the target layer is extended.
\end{enumerate}
The cut number measures robustness against arbitrary patterns of vertex
non-realization in the stipulated path graph. It is a purely combinatorial
quantity.
\end{proposition}

\begin{proof}
(a) By the reachability description, $L^{\mathrm{re}}_n(\gamma_0)$ is empty
exactly when no path from $(0,\gamma_0)$ to layer $n$ consists of realized
vertices, that is, when every such path contains an unrealized vertex
($(0,\gamma_0)$ itself is realized, since $C^{\mathrm{re}}_0=C(x_0)$). If
$L^{\mathrm{re}}_i$ is empty for some $i$ then so is $L^{\mathrm{re}}_n$,
because $R(\varnothing)=\varnothing$; so $\mathrm{Dead}^{\mathrm{re}}$ is
equivalent to $L^{\mathrm{re}}_n=\varnothing$.

(b) Add a vertex $\top$ with an edge from every layer-$n$ vertex. Since $n\ge1$,
$(0,\gamma_0)$ and $\top$ are not adjacent. A set of vertices other than
$(0,\gamma_0)$ and $\top$ meets every path from $(0,\gamma_0)$ to layer $n$ if and
only if its removal separates $(0,\gamma_0)$ from $\top$; such a set may contain
layer-$n$ vertices. Menger's theorem in its vertex form \cite{menger1927,diestel}
states that, in a finite graph, the minimum size of such a separating set equals
the maximum number of internally vertex-disjoint paths from $(0,\gamma_0)$ to
$\top$, and these correspond to paths to layer $n$ that are disjoint except at
$(0,\gamma_0)$.

(c) Since $\gamma_0$ is not stipulated-dead, every $L_i(\gamma_0)$ is
nonempty, so a path to layer $n$ exists and $\kappa_n\ge1$. For
$1\le i\le n$, every path from $(0,\gamma_0)$ to layer $n$ contains exactly one
vertex of layer $i$, so the set $\{i\}\times L_i(\gamma_0)$ meets every such
path, and $\kappa_n\le|L_i(\gamma_0)|$.

(d) Take a family of internally vertex-disjoint paths from $(0,\gamma_0)$ to
layer $n$ achieving the maximum size $\kappa_n$ by part~(b). Truncate each
path at its first vertex in layer $m$. The truncated paths are internally
vertex-disjoint, and each truncated path reaches layer $m$. A path to layer $m$
cannot always be extended to layer $n$, but a path to layer $n$ can always be
truncated at layer $m$. Therefore the maximum number of disjoint paths to layer
$m$ is at least that to layer $n$, so $\kappa_m\ge\kappa_n$ by part~(b).
\end{proof}

\begin{remark}[The bound need not be an equality]
\label{rem:bound-not-equality}
The bound in part~(c) need not be an equality. A layer may contain multiple
vertices while every path from the root to the terminal layer passes through
the same single vertex of that layer. Thus, layer cardinality gives an upper
bound on the realization-cut number, but does not by itself determine that
number.
\end{remark}

\begin{example}[Same lift sizes, different cut numbers]
\label{ex:cut-separation}
Let $\rho=(f_1,f_2,f_3)$ pass through $x_0,\dots,x_3$, with $C(x_0)=\{\gamma_0\}$
and $C(x_i)=\{u_i,v_i\}$ for $i=1,2,3$, in two systems $H_{\mathrm{par}}$ and $H_{\mathrm{fun}}$ that differ
only in their relations. In both, $R_{f_1}(\gamma_0)=\{u_1,v_1\}$. In the
\emph{parallel} system $H_{\mathrm{par}}$,
\[
R_{f_2}(u_1)=\{u_2\},\quad R_{f_2}(v_1)=\{v_2\},\quad
R_{f_3}(u_2)=\{u_3\},\quad R_{f_3}(v_2)=\{v_3\}.
\]
In the \emph{funnel} system $H_{\mathrm{fun}}$,
\[
R_{f_2}(u_1)=\{u_2\},\quad R_{f_2}(v_1)=\{u_2,v_2\},\quad
R_{f_3}(u_2)=\{u_3,v_3\},\quad R_{f_3}(v_2)=\varnothing .
\]
In both systems $|L_i(\gamma_0)|=1,2,2,2$ for $i=0,1,2,3$, so $\gamma_0$ is
diluted along $\rho$ with the same cardinality profile and no empty lift. The
persistence statuses and the layer sizes do not separate the two. The cut number
does. In $H_{\mathrm{par}}$ the chains $u_1u_2u_3$ and $v_1v_2v_3$ are vertex-disjoint, so
$\kappa_3(\gamma_0)=2$. In $H_{\mathrm{fun}}$ the continuation $v_2$ belongs to the lift but has
no successor, so every path to layer $3$ passes through $(2,u_2)$, and
$\kappa_3(\gamma_0)=1$.

The difference is operational in the following sense. Suppose the determination
process can leave at most one vertex unrealized, in any pattern. In $H_{\mathrm{par}}$ the
realized lift of $\gamma_0$ is nonempty whichever single vertex fails. In $H_{\mathrm{fun}}$ the
single failure of $(2,u_2)$ empties it, while the failure of $(2,v_2)$ or of
either layer-1 vertex is harmless. A minimum cut also says \emph{where} the
fragility sits: here it is the vertex $(2,u_2)$, which is where redundancy would
have to be added, or where realization would have to be verified most carefully,
to raise $\kappa_3$. This is a worst-case statement about a bounded number of
failures, in the sense of the following remark, and not a probability, and it
concerns reaching the terminal layer, not whether an excluded coordinate can be
corrected.
\end{example}

\begin{remark}[Operational meaning of the cut number]
The cut number measures robustness against arbitrary patterns of vertex
non-realization. The conformance-determination process may produce only some of
these patterns. A determination-constrained version would require a separate
specification of which failure patterns the realization layer permits. The cut
number treats vertex non-realizations as combinatorially available failure
events and does not model correlations among them; a constrained robustness
measure would require an additional specification of which failure sets the
realization process permits.
The cut number is a structural quantity of the fully realized path graph.
Reading it as tolerance to failure requires an additional assumption on the
determination process: a bound on how many vertices it can leave unrealized. If
at most $\kappa_n-1$ vertices can be unrealized, in any pattern and with any
correlation, the lift cannot be emptied; no independence assumption is needed
for this worst-case guarantee, and independence matters only for probabilistic
statements. Without such a bound, a large cut number does not by itself
guarantee that a real determination process fails to empty the lift.
\end{remark}

\begin{remark}[Computational status of the cut number]
Once the path $\rho$ is fixed, the path graph $G_\rho(\gamma_0)$ has size
polynomial in the extensional input, and $\kappa_n(\gamma_0)$ is computable in
polynomial time by max-flow with vertex splitting (equivalently, by the standard
algorithmic proof of Menger's theorem). This contrasts with the unrestricted
death-existence problem of Appendix~\ref{app:complexity-proofs}, which is
PSPACE-hard: hardness enters in searching over which path to take, not in
evaluating robustness along a fixed path.
\end{remark}

\begin{remark}[Relation to recoverability]
\label{rem:cut-recoverability}
The cut number and the recoverability predicates of
Section~\ref{sec:recoverability} are read off the same lifts, and Menger's
theorem applies to both. Work under the realized definitions of
Appendix~\ref{app:realization}, with finite fibers, and let $\rho'$ be a
generating history of length $n$ together with an admissible recognition
continuation.

\emph{Lineage death.} Let $\gamma_0\in I_0$. If fewer than $\kappa_n(\gamma_0)$
vertices of $G_{\rho'}(\gamma_0)$ are unrealized, the set of unrealized vertices
is not a cut, so $L^{\mathrm{re}}_n(\gamma_0)\neq\varnothing$ by
Proposition~\ref{prop:robustness}(a),(b). Emptiness at an earlier stage persists
to stage $n$, so no lineage-death event for $\gamma_0$ is generated at any prefix
of $\rho'$. If this holds for every $\gamma_0\in I_0$, the requirement of
$\operatorname{Recover}_{\mathrm{avail}}$ that $F(\rho')$ contain no lineage-death
event is met. The condition is sufficient, not necessary.

\emph{Corrective routes.} Let $e=(\rho_e,\gamma)$ be an unresolved exclusion event
at state $x$, and let $G$ be the path graph of the segment of $\rho'$ from $x$,
with root $(0,\gamma)$ and vertices $(i,\delta)$, $\delta$ in the stipulated lift
of $\gamma$ after $i$ steps. Call $(i,\delta)$ with $i\ge1$ a \emph{witness
vertex} if $(\eta,\delta)$ is a corrective-route witness for $e$ on $\rho'$,
where $\eta$ is the prefix of $\rho'$ ending at layer $i$. Add a sink $\top$ with
an edge from every witness vertex. The root is not a witness vertex, so it is not
adjacent to $\top$, and Menger's theorem gives: the least number of vertices other
than the root and $\top$ whose removal separates the root from $\top$ equals the
maximum number of root--$\top$ paths that are pairwise vertex-disjoint except at
the root and $\top$. A non-realized vertex is deleted, since it cannot be a
witness, so no set of non-realizations smaller than this number removes every
witness for $e$ on $\rho'$.

So the cut number measures how robust the existence conditions inside
$\operatorname{Recover}_{\mathrm{avail}}$ are to partial realization; it does not
enter the recognition conditions $B_\eta(y)(\delta)=\mathsf U$ and
$\widehat S(y,\delta)=\mathsf I$. Along a given extension both are computed by
maximum flow in polynomial time, so the hardness of
Proposition~\ref{prop:recover-avail-classification} lies in choosing the
extension and in requiring one extension to serve every unresolved event.
\end{remark}

\begin{remark}[The cut number is path-indexed]
\label{rem:cut-path-indexed}
The cut number $\kappa_n(\gamma_0)$ is defined relative to a fixed realized
history $\rho$. Two different histories reaching the same state may traverse
different modifications and carry different stipulated lifts, and the cut
numbers along them are separate quantities. The framework does not define an
aggregate robustness measure over a set of paths. A natural definition would
take the minimum of $\kappa_n$ over an admissible set of histories, or the
minimum over histories that respect the determination process, but the choice
of admissible set is a governance act of the same kind as the choice of $D$.
Aggregation across paths, and the choice of an aggregate, are not treated
here.
\end{remark}

The cut number concerns reaching the terminal layer. Whether a particular
excluded coordinate can be corrected also depends on which terminal vertices
descend from it, lie in $D$, and are admitted by $S$, so a lineage with a large
$\kappa_n$ may still have no corrective route.

A preserved lineage has $|L_i|=1$ at every stage, hence $\kappa_n=1$ by~(c): a
single unrealized link kills it. A diluted lineage has
$\kappa_n\le\min_i|L_i|$, so a singleton layer anywhere in layers $1$ through
$n$ caps $\kappa_n$ at $1$, however wide the other layers are. The statuses
classify structural branching, whereas $\kappa_n$ counts how many independent
realization failures can be tolerated; the two are different quantities.
Conversely, $\kappa_n\ge2$ requires that $\gamma_0$ be diluted along the
prefix: a preserved lineage has a unique path to the terminal layer and cannot
have more than one independent path.

In the abstract model, the stipulated lift of $\gamma_0$ along $P_A$ has
cardinalities $1,1,1,2,2,2$ at $x_0,x_1,x_2,x_4,x_5,x_6$. By~(c),
$\kappa_5(\gamma_0)\le|L_1(\gamma_0)|=1$, so $\gamma_0$ is diluted but exactly
as fragile as a preserved lineage: leaving $\gamma_1$ or $\gamma_2$ unrealized
kills it. For the lineage of $\gamma_2$ alone along $(d,f,g)$, the $S$-chain
$\gamma_4^S,\gamma_5^S,\gamma_6^S$ and the $W$-chain
$\gamma_4^W,\gamma_5^W,\gamma_6^W$ are vertex-disjoint paths, so $\kappa=2$.
Survival is not the same as every layer being nonempty: realizing
$\gamma_4^S$ and $\gamma_5^W$ but nothing else on the fork empties the lineage,
because there is no edge between the chains.

\subsection{Determinations}

A conformance determination is recorded as
\[
\mathrm{det}_i=(x_i,\ \rho|_i,\ C^{\mathrm{re}}_i,\ \mathrm{ev}_i,\ \mathrm{dec}_i),
\]
where $\mathrm{ev}_i$ is the evidence on which it rests and $\mathrm{dec}_i$ identifies the decider.
The particular schema is illustrative; what matters is that the determination
record separates realization from stipulation and recognition, and that it
carries the evidence and the decider so that the judgment is attributable. A
determination assigns each continuation in $C(x_i)$ to the realized subset
$C^{\mathrm{re}}_i$ or its complement. The assignment may incorporate
evidential uncertainty; the framework records the determination rather than
prescribing how uncertainty is resolved. Whether the determination was correct
is outside the formalism.

Whether a symmetry-breaking determination reflects system-side evidence or
normative supplementation is a classification of its recorded justification,
not a property visible to the augmented structure
$(H,C,R,C^{\mathrm{re}}_\rho)$.

\subsection{Underdetermination with determinations}

\begin{remark}[Underdetermination with determinations]
\label{rem:realized-symmetry}
Let $\sigma$ be a descendant-exchanging symmetry as in
Corollary~\ref{cor:underdetermination}, exchanging $\delta_i$ and $\delta_j$
at a modification $f:x_m\to x_{m+1}$ with $D$ invariant under $\sigma$, and let
$\rho$ be a finite realized history through $f$ that $\sigma$ fixes pointwise.
Call an automorphism of $(H,C,R)$ an automorphism of the
\emph{augmented structure} $(H,C,R,C^{\mathrm{re}}_\rho)$ if it maps each
$C^{\mathrm{re}}_l$ onto itself. If the determinations realize both or neither
of $\delta_i,\delta_j$ and are otherwise invariant under $\sigma$, then
$\sigma$ is an automorphism of the augmented structure, the realized decision
domain is invariant, and the argument for
Corollary~\ref{cor:underdetermination} applies unchanged: the determination
has not broken the symmetry. If the determination realizes exactly one of them,
$\sigma$ is not an automorphism of the augmented structure, and the asymmetry
between $\delta_i$ and $\delta_j$ has been introduced by the determination and
not by the stipulated structure. A symmetry-breaking determination is
classified, for the purposes of the record, as a normative supplementation when
its recorded justification supplies no system-side evidential distinction
between the exchanged descendants; this is a classification of the record, not
a claim about the underlying system. In the abstract model, a determination at
$x_4$ along $P_A$ that realizes $\gamma_4^S$ but not $\gamma_4^W$ shrinks the
subgroup generated by $\sigma$ from order two to trivial: $\gamma$ is diluted
in the stipulated structure and preserved in the realized one. The recognition
record then no longer requires a choice between $\gamma_4^S$ and $\gamma_4^W$
on $P_A$, because only one of them is admitted by the realized decision domain.
\end{remark}

\subsection{Effect on recoverability}

The main text uses the full-realization baseline $D^{\mathrm{re}}_\rho=D$.
Under partial realization, the same definitions use realized lifts and domains.
At stage $i$ of a realized history $\rho$, with reached state $x_i$, the
realized decision domain is
$D^{\mathrm{re}}_{\rho_i}(x_i)=D(x_i)\cap C^{\mathrm{re}}_i$ and the
recognition state is $B_{\rho_i}(x_i)$. An exclusion event is a pair
$(\rho_e,\gamma)$ with
\[
\gamma\in D^{\mathrm{re}}_{\rho_e}(x),
\qquad
B_{\rho_e}(x)(\gamma)=\mathsf E,
\]
where $\rho_e$ is a finite realized history ending at $x$ and the recognition
state is $B_{\rho_e}$. A lineage-death event is a pair $(\rho_e,\gamma_0)$ with
$\gamma_0\in I_0$, where $\rho_e$ ends at the first stage at which the realized
lift $L^{\mathrm{re}}_i(\gamma_0)$ is empty. For a path segment $p$ along a
realized history, write $R^{\mathrm{re}}_p(\gamma)$ for the corresponding
recursively realized lift, with the realized subset intersected at each
intermediate stage. A corrective inscription for an exclusion event at a later
state $y$, reached at stage $j$ along an extension, may inscribe a descendant
$\delta$ satisfying
\[
\delta\in R^{\mathrm{re}}_p(\gamma)\cap D^{\mathrm{re}}_{\rho_j}(y),
\qquad
B_{\rho_j}(y)(\delta)=\mathsf U,
\qquad
S(y,\delta)=\mathrm{inscribed},
\]
followed by the admissible transition
\[
B_{\rho_{j+1}}(y)(\delta)=\mathsf I.
\]
Thus realization supplies eligibility, $S$ supplies recognition admissibility,
and the recognition history records occurrence. The criterion $S$ is defined
on all of $C$, but in the realized setting its value at a coordinate outside
$D^{\mathrm{re}}_{\rho_j}(y)$ is not consulted: the realized decision domain
is $D^{\mathrm{re}}_{\rho_j}(y) = D(y) \cap C^{\mathrm{re}}_j$, and a
coordinate excluded from it by non-realization receives no decision.
A realized lineage-death event
is attached to the realized history on which it is generated, and admits no
corrective inscription along extensions of that history: on any extension of
$\rho_e$, the realized lift stays empty by the recursion
\[
L^{\mathrm{re}}_{i+1}=R_{f_{i+1}}(L^{\mathrm{re}}_i)\cap C^{\mathrm{re}}_{i+1},
\]
under which $R_{f_{i+1}}(\varnothing)=\varnothing$, not by a composition law
for the realized relation. The qualification is local to $\rho_e$: a different
history with a different determination may have a nonempty realized lift for
the same stipulated lineage.

The determination process remains fixed across quantified extensions; later
outcomes may still affect whether a required descendant is realized. The
determination record identifies the decider responsible for those outcomes.

\section{Detailed Calculations for the Abstract Model}
\label{app:calc}

For the A branch,
\begin{equation}
R_d(R_b(\gamma_1))
=
R_d(\{\gamma_2\})
=
\{\gamma_4^S,\gamma_4^W\}.
\end{equation}

For the B branch, the stipulated lift of $\gamma_0$ becomes empty at $c$, so
\begin{equation}
R_c(R_a(\gamma_0))
=
\varnothing.
\end{equation}

$C(x_4)$ is the continuation space available at $x_4$; the reachable subset
along $P_A$ is $\{\alpha_4,\beta_4,\gamma_4^S,\gamma_4^W\}$, and along $P_B$ is
$\{\alpha_4,\beta_4\}$.

For $\alpha$ and $\beta$, every local lift has cardinality one, so
\[
\mathrm{status}_{P_A}(\alpha_0)
=
\mathrm{status}_{P_B}(\alpha_0)
=
\mathrm{Pres},
\qquad
\mathrm{status}_{P_A}(\beta_0)
=
\mathrm{status}_{P_B}(\beta_0)
=
\mathrm{Pres}.
\]

For $\gamma$, $|R_d(R_b(\gamma_1))|=2$, so
$\mathrm{status}_{P_A}(\gamma_0)=\mathrm{Dil}$,
whereas $R_c(R_a(\gamma_0))=\varnothing$ and therefore
$\mathrm{status}_{P_B}(\gamma_0)=\mathrm{Dead}$, both under the stipulated
structure.

The nonempty generating relations used in the example are
\begin{align*}
R_a(\alpha_0)&=\{\alpha_1\}, & R_a(\beta_0)&=\{\beta_1\}, & R_a(\gamma_0)&=\{\gamma_1\}, \\
R_b(\alpha_1)&=\{\alpha_2\}, & R_b(\beta_1)&=\{\beta_2\}, & R_b(\gamma_1)&=\{\gamma_2\}, \\
R_c(\alpha_1)&=\{\alpha_3\}, & R_c(\beta_1)&=\{\beta_3\}, & R_c(\gamma_1)&=\varnothing, \\
R_d(\alpha_2)&=\{\alpha_4\}, & R_d(\beta_2)&=\{\beta_4\}, & R_d(\gamma_2)&=\{\gamma_4^S,\gamma_4^W\}, \\
R_e(\alpha_3)&=\{\alpha_4\}, & R_e(\beta_3)&=\{\beta_4\}, & R_e(\gamma_3)&=\varnothing, \\
R_f(\alpha_4)&=\{\alpha_5\}, & R_f(\beta_4)&=\{\beta_5\}, & R_f(\gamma_4^S)&=\{\gamma_5^S\}, & R_f(\gamma_4^W)&=\{\gamma_5^W\}, \\
R_g(\alpha_5)&=\{\alpha_6\}, & R_g(\beta_5)&=\{\beta_6\}, & R_g(\gamma_5^S)&=\{\gamma_6^S\}, & R_g(\gamma_5^W)&=\{\gamma_6^W\}.
\end{align*}
The displayed relations, including the explicitly empty relations, exhaust the
generating compatibility relations for the core system. The displayed relations
for $\alpha$ and $\beta$ through $f$ and $g$ are stipulated as singleton
continuation relations; relations for composite modifications then follow from
the composition condition.

For the local decision domain concentrated at $x_4$, let $D(x_0)=I_0$, let
$D(x_4)=\{\gamma_4^S,\gamma_4^W\}$, and let $D(x)=\varnothing$ elsewhere in the
core system. Along $P_A$ the realized decision domain at $x_4$ is
$D^{\mathrm{re}}_{P_A}(x_4)=\{\gamma_4^S,\gamma_4^W\}$. Along $P_B$, a
conformance determination is stipulated that excludes both $\gamma_4^S$ and
$\gamma_4^W$ from $C^{\mathrm{re}}_{P_B}(x_4)$, so
$D^{\mathrm{re}}_{P_B}(x_4)=\varnothing$ by determination. Writing only the
noninitial $x_4$ component of the possible assignment, the four possibilities
are in bijection with the four elements of $2^{D(x_4)}$.

For recoverability, suppose a finite prefix $\rho_{10}$ contains exactly one
unresolved exclusion event $(\rho_{10}, \gamma_{10})$ at state $x_{10}$ and no
lineage-death events. A subsequent modification $r: x_{10} \to x_{13}$ with
$R_r(\gamma_{10}) = \{\gamma_{13}^S\}$, where $\gamma_{13}^S$ is a decision
coordinate not yet adjudicated, then provides a corrective route for the unique
unresolved exclusion event. Since there are no lineage-death events and no
other unresolved exclusions, the existence of this route establishes
$\operatorname{Recover}_{\mathrm{avail}}(\rho_{10})$ for this example. The
prefix $\rho_{10}$ alone does not establish this.

If instead $L_i(\gamma_0)=\varnothing$ at some stage $i$ of $\rho_{10}$ under
the stipulated structure, the resulting lineage-death event would remain
unresolved on every extension of that prefix, and
$\operatorname{Recover}_{\mathrm{avail}}(\rho_{10})$ would fail regardless of
any corrective route for the exclusion event.

\subsection{The automorphism group of the core system}
\label{app:aut-toy}

Let $\sigma\in\Aut_*$ be a symmetry of the pointed structure
$(H,C,R,x_0,I_0,D)$ of Section~\ref{sec:model}. We verify that
$\Aut_*=\mathbb{Z}/2\times\mathbb{Z}/2$.

\paragraph{States are fixed.} By definition $\sigma(x_0)=x_0$. The only
modification out of $x_0$ is $a$, so $\sigma(a)=a$, hence $\sigma(x_1)=x_1$.
From $x_1$ the modifications $b:x_1\to x_2$ and $c:x_1\to x_3$ are the only
ones, so $\sigma$ permutes $\{b,c\}$, hence $\{x_2,x_3\}$. Were $\sigma$ to
swap $b$ and $c$, we would have $\sigma(x_2)=x_3$ and $\sigma(x_3)=x_2$, and
so $\sigma$ would map the fiber $C(x_2)$ onto $C(x_3)$. Then
$\sigma(R_d(\gamma_2)) = R_e(\sigma(\gamma_2))$ is required, but
$|R_d(\gamma_2)|=2$ while every relation out of $C(x_3)$ has image of size at
most $1$ ($R_e(\alpha_3)=\{\alpha_4\}$, $R_e(\beta_3)=\{\beta_4\}$,
$R_e(\gamma_3)=\varnothing$). Contradiction. So $\sigma$ fixes $b$, $c$,
$x_2$, $x_3$ individually, and then fixes $d$, $e$, $x_4$, and, by the same
argument on $f,g$, the states $x_5,x_6$.

\paragraph{The $\gamma$-chain up to $x_3$ is fixed.} Suppose
$\sigma(\gamma_0)=\alpha_0$. Then equivariance of $R_a$ gives
$\sigma(\gamma_1)=\alpha_1$, of $R_b$ gives $\sigma(\gamma_2)=\alpha_2$, and
of $R_d$ gives $\sigma(R_d(\gamma_2))=R_d(\alpha_2)$. But $|R_d(\gamma_2)|=2$
while $|R_d(\alpha_2)|=1$, and $\sigma$ is a bijection, so no such $\sigma$
exists. The same argument excludes $\sigma(\gamma_0)=\beta_0$. Hence $\sigma$
fixes $\gamma_0$, and then fixes $\gamma_1,\gamma_2$ by the same equivariance.
It also fixes $\gamma_3$: equivariance gives
$R_e(\sigma\gamma_3)=\sigma(R_e(\gamma_3))=\varnothing$, while $R_e(\alpha_3)$ and
$R_e(\beta_3)$ are nonempty.

\paragraph{The $\gamma$-fork at $x_4$ is fixed setwise.} The constraint
$\sigma(D(x_4))=D(x_4)$ with $D(x_4)=\{\gamma_4^S,\gamma_4^W\}$ forces
$\sigma(\{\gamma_4^S,\gamma_4^W\})=\{\gamma_4^S,\gamma_4^W\}$. Combined with
$R_f$ and $R_g$ (both singletons), this gives the same for
$\{\gamma_5^S,\gamma_5^W\}$ and $\{\gamma_6^S,\gamma_6^W\}$, and equivariance of
$R_f$ and $R_g$ ties the exchange at $x_5$ and $x_6$ to the exchange at $x_4$:
$\sigma$ swaps all three pairs or none.

\paragraph{The $\alpha/\beta$ layer is fixed or swapped.} Since $\sigma$ fixes
$\gamma_0$ and permutes $\{\alpha_0,\beta_0\}$, and the relations $R_a,R_b,R_c,
R_d,R_e,R_f,R_g$ act consistently ($\alpha\leftrightarrow\beta$ throughout),
either $\sigma$ fixes $\alpha_i,\beta_i$ for all $i$ or it swaps them for all
$i$.

\paragraph{Generators.} Let $\sigma_{\alpha\beta}$ swap $\alpha_i\leftrightarrow
\beta_i$ for all $i$ and fix all $\gamma$-coordinates. Let $\sigma_\gamma$ swap
$\gamma_4^S\leftrightarrow\gamma_4^W$, $\gamma_5^S\leftrightarrow\gamma_5^W$,
$\gamma_6^S\leftrightarrow\gamma_6^W$ and fix all $\alpha$- and
$\beta$-coordinates. Both are automorphisms of the pointed structure: each
preserves $x_0$, $I_0$ (as a set), and $D(x_4)$ (as a set), and each commutes
with the relations. They commute with each other, and the paragraph above shows
that every element of $\Aut_*$ is one of
$\{\mathrm{id},\sigma_{\alpha\beta},\sigma_\gamma,\sigma_{\alpha\beta}\sigma_\gamma\}$.

Hence $\Aut_*=\mathbb{Z}/2\times\mathbb{Z}/2$, of order four. Its orbits on
the coordinates of $D$ with $x\neq x_0$ are $\{\gamma_4^S,\gamma_4^W\}$,
$\{\gamma_5^S,\gamma_5^W\}$, $\{\gamma_6^S,\gamma_6^W\}$, together with the
singleton orbits of the $\alpha$- and $\beta$-coordinates that lie in $D$
(none, in the core system). The characterization of
Section~\ref{sec:underdetermination} then applies directly: the only target set
in $D$ at a state $y\neq x_0$ is $\{\gamma_4^S,\gamma_4^W\}$ at $x_4$, its
orbit meets it in two elements, and so no structural exclusive criterion
exists.

\section{Notation}\label{app:notation}
Symbols used throughout the paper are collected below, grouped by topic; the last column gives the section in which each is introduced. Symbols that occur only inside a single proof or construction (for example, the clause continuations $\xi^{(i)}_j$, the variables $\nu_i$ and the NFA transition function $\mu$ in Appendices~\ref{app:complexity-proofs} and~\ref{app:recover-avail-verification}) are defined where they are used and are not listed.

\begin{longtable}{@{}>{\raggedright\arraybackslash}p{0.27\textwidth}>{\raggedright\arraybackslash}p{0.60\textwidth}r@{}}
\hline Symbol & Meaning & Section \\ \hline \endhead
\multicolumn{3}{@{}l}{\textit{Structure}} \\
$H$, $X$ & directed state/modification structure; its set of states & \ref{sec:continuation} \\
$x_0$ & initial state & \ref{sec:persistence} \\
$C(x)$ & continuation space at state $x$ & \ref{sec:continuation} \\
$\gamma,\ \delta,\ \varepsilon$ & elements of continuation spaces: $\gamma_0\in C(x_0)$ is the tracked continuation, $\delta$ a descendant, $\varepsilon$ a generic element of a lift & \ref{sec:continuation} \\
$R_f$ & compatibility relation $R_f\subseteq C(x)\times C(y)$ of a modification $f:x\to y$ & \ref{sec:continuation} \\
$R_f(\gamma)$, $R_f(\Gamma)$ & lift of a continuation, respectively of a set $\Gamma\subseteq C(x)$ & \ref{sec:continuation} \\
$R_p$ & relation of a path $p$ (the relation of its composite) & \ref{sec:persistence} \\
\multicolumn{3}{@{}l}{\textit{Persistence}} \\
$L_i(\gamma_0)$ & iterated lift after $i$ steps along a path & \ref{sec:persistence} \\
$\mathrm{Dead}_p,\ \mathrm{Pres}_p,\ \mathrm{Dil}_p$ & persistence statuses of $\gamma_0$ along a path $p$ & \ref{sec:persistence} \\
$\mathrm{status}_p(\gamma_0)$ & the unique status of $\gamma_0$ along $p$ & \ref{sec:persistence} \\
\multicolumn{3}{@{}l}{\textit{Recognition}} \\
$I_0$ & initial inscription at $x_0$ & \ref{sec:decision} \\
$D(x)$, $\Dec$ & decision domain at $x$; disjoint union of decision coordinates & \ref{sec:decision} \\
$I(x)$ & completed recognition record at $x$ & \ref{sec:decision} \\
$\mathscr{R}_H(I_0,D)$ & space of possible recognition assignments & \ref{sec:decision} \\
$\rho,\ \rho_i,\ \rho'$ & finite history, its prefixes, and an extension of it & \ref{sec:recognition-histories} \\
$\tau,\ \tau|i$ & infinite history and its length-$i$ prefix & \ref{sec:recognition-histories} \\
$\preceq$ & prefix order on histories & \ref{sec:recognition-histories} \\
$B_\rho$ & recognition state at prefix $\rho$, with values $\mathsf I,\mathsf E,\mathsf U$ (inscribed, excluded, undecided) & \ref{sec:recognition-histories} \\
$S$ & recognition criterion (static map) & \ref{sec:underdetermination} \\
$\widehat S$ & partial binary admissibility map & \ref{sec:underdetermination} \\
\multicolumn{3}{@{}l}{\textit{Symmetry and underdetermination}} \\
$\Aut(H,C,R)$ & group of structural symmetries & \ref{sec:underdetermination} \\
$\Aut_*$, $\sigma$ & pointed automorphism group and an element of it & \ref{sec:underdetermination} \\
$T$ & target set: the lift of a continuation in $I_0$ along a path, intersected with $D(y)$ & \ref{sec:underdetermination} \\
$\mathcal O$ & set of $\Aut_*$-orbits & \ref{sec:underdetermination} \\
$P_A,\ P_B$ & the two maximal generating histories of the core system & \ref{sec:model} \\
$I_C,\ I_D,\ I_B,\ I_W$ & the four possible assignments at $x_4$ (conservative, double, blotted, weak) & \ref{sec:model} \\
\multicolumn{3}{@{}l}{\textit{Recoverability}} \\
$F(\rho)$ & failure events generated along $\rho$ & \ref{sec:recoverability} \\
$U(\rho)$ & unresolved failure events & \ref{sec:recoverability} \\
$W_e(\rho')$ & witness set for a failure event $e$ on $\rho'$ & \ref{sec:recoverability} \\
$\mathrm{Recover}_{\mathrm{avail}}$, $\mathrm{Recover}_{\mathrm{trace}}$ & availability and trace recoverability & \ref{sec:recoverability} \\
$\mathcal P$ & set of infinite histories satisfying trace recoverability & \ref{sec:recoverability} \\
\multicolumn{3}{@{}l}{\textit{Realization}} \\
$\Delta$ & determination rule & \ref{sec:realization-summary} \\
$C^{\mathrm{re}}_i$, $D^{\mathrm{re}}$ & realized subset at stage $i$; realized decision domain & \ref{sec:realization-summary} \\
$L^{\mathrm{re}}_i$, $R^{\mathrm{re}}_p$ & realized lift at stage $i$; realized relation of a path & \ref{app:realization} \\
$\mathrm{Dead}^{\mathrm{re}},\ \mathrm{Pres}^{\mathrm{re}},\ \mathrm{Dil}^{\mathrm{re}}$ & realized statuses & \ref{app:realization} \\
$G_\rho(\gamma_0)$ & path graph of a lineage along $\rho$ & \ref{sec:realization-summary} \\
$\kappa_n(\gamma_0)$ & realization-cut number along a path & \ref{sec:realization-summary} \\
\hline
\end{longtable}

\end{document}